\documentclass[conference]{IEEEtran}
\IEEEoverridecommandlockouts
\usepackage{cite}
\usepackage{amsmath,amssymb,amsfonts}
\usepackage{algorithmic}
\usepackage{graphicx}
\usepackage{textcomp}
\usepackage{xcolor}

\usepackage{subcaption}
\usepackage{makecell}
\usepackage{multirow}
\usepackage{mathtools}
\usepackage{algorithm}
\usepackage{algorithmic}
\usepackage{graphicx}
\usepackage{booktabs}
\usepackage{hyperref}

\def\BibTeX{{\rm B\kern-.05em{\sc i\kern-.025em b}\kern-.08em
    T\kern-.1667em\lower.7ex\hbox{E}\kern-.125emX}}
\begin{document}
\newcommand{\update}[1]{\textcolor{red}{#1}}
\title{ReasonCache: Accelerating Large Reasoning Model Serving through KV Cache Sharing}

\newcommand{\tx}[1]{{\color{blue} TX: #1}}

\author{
    %Authors
    % All authors must be in the same font size and format.
    Kaiwen Chen\textsuperscript{\rm 1},
    Xin Tan\textsuperscript{\rm 1},
    Minchen Yu\textsuperscript{\rm 2},
    Jingzong Li\textsuperscript{\rm 3},
    Hong Xu\textsuperscript{\rm 1}
    \\
    \textsuperscript{\rm 1}The Chinese University of Hong Kong\\
    \textsuperscript{\rm 2}The Chinese University of Hong Kong, Shenzhen\\
    \textsuperscript{\rm 3}The Hang Seng University of Hong Kong\\
}

\maketitle

\begin{abstract}
    Large Reasoning Models (LRMs) are becoming integral to many AI inference systems, enhancing their capabilities with advanced reasoning.
    However, deploying these models in production environments presents a significant QoS challenge: the substantial memory overhead from their long, auto-regressive inference processes severely limits throughput and increases latency, thereby affecting the quality of service for concurrent users.
    We observe that LRMs frequently generate highly similar intermediate reasoning steps, which, in turn, correspond to highly similar KV cache states across layers. 
    Building on this insight, we propose ReasonCache, a novel KV cache management approach designed to improve the QoS of AI inference systems.
    ReasonCache utilizes a Collaborative Filtering Algorithm to efficiently identify reusable KV cache blocks and enables zero-copy cache reuse.
    Experimental evaluation demonstrates that ReasonCache achieves a peak throughput improvement of 89.2\% and an average gain of 40-60\%, leading to more responsive and cost-effective AI inference services. 
    Notably, this performance is achieved while maintaining higher accuracy compared to existing KV cache management techniques.
\end{abstract}


\section{Introduction}
The deployment of Large Reasoning Models (LRMs), 
such as OpenAI's o1~\cite{openai-o1}, QwQ-32B~\cite{QwQ} and DeepSeek-R1~\cite{deepseek-r1}, 
is becoming critical for AI inference systems. 
These models are increasingly powering core features in applications like search engines~\cite{reasonrank}, recommender systems~\cite{reasontorecommend}, and conversational AI agents~\cite{agent}, delivering sophisticated reasoning capabilities that enhance user satisfaction. 
To achieve this, LRMs perform extensive reasoning before producing a final answer, including long chain-of-thought (CoT) sequences, exploration of multiple strategies, fine-grained problem decomposition, and self-verification ~\cite{LRMS,overthinking}.

\begin{figure}[t]
    \centering
    \includegraphics[width=\columnwidth]{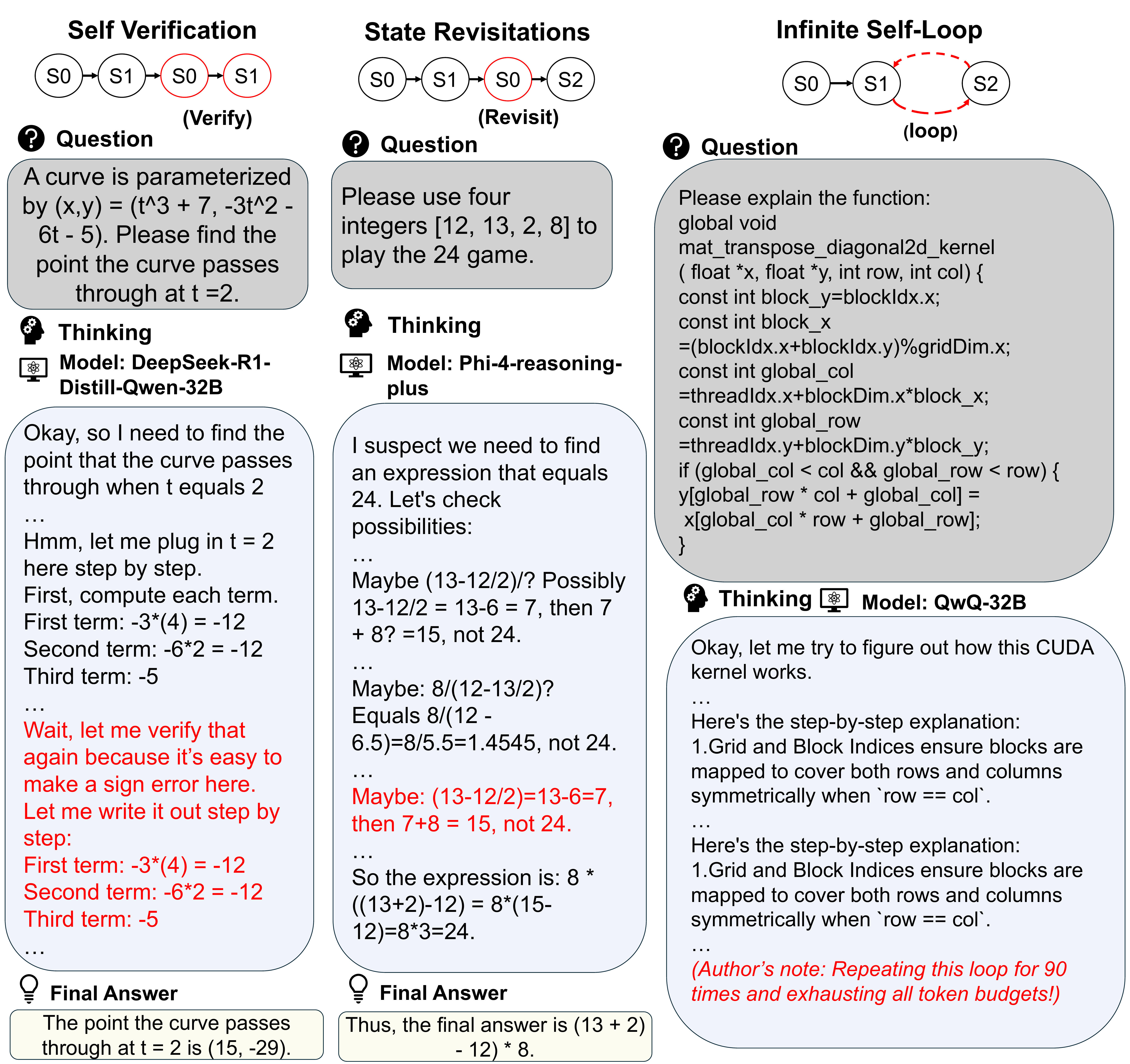}
    \caption{Examples of \textit{redundant thinking} patterns in Large Reasoning Models (LRMs). }
    \label{fig1}
    \vspace{-4mm}
  \end{figure}

Meanwhile, Key-value (KV) caching accelerates autoregressive generation in LRMs by storing the attention states of preceding tokens.
While reusing these states avoids redundant computation and enhances inference speed, it incurs significant memory overhead. 
This overhead is particularly pronounced in LRMs due to their extensive reasoning processes.
For instance, recent research~\cite{overthinking} shows that the current LRM, QwQ-32B, can generate up to 13 redundant solutions for a trivial query such as ``What is 2 + 3?'', 
resulting in a 1,953\% increase in token count compared to conventional models. 
On the AIME~24~\cite{aime} dataset, the model produces an average of 12,406 tokens per problem \footnote{Tested with Qwen team's recommended sampling parameters: $Temperature=0.6, TopP=0.95, TopK=20$.}. 
With each token occupying 1280 KB of memory\footnote{Calculated as 2 (key/value vectors) $\times$ 5120 (hidden dimensions) $\times$ 64 (number of layers) $\times$ 2 (FP16 size). }, 
the KV cache alone requires roughly 15 GB of GPU memory per request. 
Given the typical memory constraints of modern GPUs, such substantial consumption severely limits the system's maximum concurrency.

Recent advances in KV cache management primarily exploit attention sparsity~\cite{sparse-transformer} to reduce memory overhead. 
These methods can be categorized into three types: KV cache eviction~\cite{h2o,streaming-llm,snapkv,lethe,raas,RLKV}, selective loading~\cite{quest}, and KV cache merging~\cite{Minicache,d2o}. 
Their core idea is to estimate the importance of tokens based on attention sparsity and then selectively discard, load, or merge tokens accordingly. 
However, these methods suffer from inherent limitations:
eviction is irreversible, risking the permanent loss of tokens that may become important later; 
selective loading fails to reduce the memory footprint; 
and merging can cause hallucinations due to inaccurate importance prediction.
Furthermore, a critical limitation common to all these sparsity-based techniques is their fundamental incompatibility with PagedAttention~\cite{vllm,sglang}. 
Exploiting attention sparsity requires dynamic, fine-grained operations, as sparsity patterns vary across layers and the set of critical tokens is query-dependent~\cite{Minicache,quest,dsv}. 
Consequently, these methods resort to token- or layer-level manipulations, which disrupt the contiguous block-based memory management of modern serving systems~\cite{vllm,sglang}, and lead to memory fragmentation.

Instead of focusing on attention sparsity, our work identifies a novel optimization opportunity. 
We observe that LRMs frequently engage in what we term \textit{redundant thinking}: the generation of repetitive reasoning steps with highly similar content during problem-solving.
This behavior manifests in distinct patterns such as self-verification, state revisitation, and infinite self-loops, as illustrated in Figure \ref{fig1}. 
More importantly, these similar steps exhibit highly similar KV cache states. Key and value vectors from similar reasoning steps remain closely aligned across transformer layers.
This insight enables a clear optimization opportunity: by sharing similar KV cache blocks, we can reduce memory overhead and enhance inference efficiency without compromising accuracy.

However, realizing this opportunity requires overcoming two significant technical challenges. 
First, reusable KV cache blocks must be identified accurately and efficiently. 
A naive pairwise comparison of all blocks is computationally infeasible due to its quadratic complexity ($O(n^2)$) with respect to sequence length. 
Second, the reuse mechanism itself must be seamlessly integrated into the decoding process without introducing additional GPU memory bandwidth overhead or disrupting its highly parallelized nature.

To address these challenges, we introduce ReasonCache, a system that efficiently identifies reusable KV cache blocks using a two-stage, coarse-to-fine filtering approach. 
To identify similar blocks, ReasonCache first performs a lightweight, step-level comparison to prune the search space.
It then employs a fine-grained, block-level similarity check, measuring the Euclidean distance between key and value vectors to accurately confirm reusability. 
For seamless integration, ReasonCache is designed at the block level, aligning with the memory management schemes of modern serving frameworks like vLLM~\cite{vllm} and SGLang~\cite{sglang}.

% To exploit this insight, we propose ReasonCache, which employs Collaborative Filtering Algorithm 
% that identifies reusable KV cache blocks through a two stage hierarchical process. 
% First, it performs lightweight step-level comparisons to narrow down candidate blocks, reducing computational overhead. 
% Second, it precisely measures block-level Euclidean distances between key and value vectors to ensure reusability. 
% Additionally, ReasonCache operates at the block-level, seamlessly aligning with vLLM and SGLang.
% This approach significantly mitigates memory overhead, increases throughput while maintaining accuracy.

To validate the effectiveness of ReasonCache, we conduct comprehensive evaluations across a range of tasks to assess its impact on the QoS.
Our experimental results demonstrate that ReasonCache delivers significant performance gains with minimal accuracy loss. 
It achieves an average throughput improvement of 40\% to 60\% compared to dense attention, while retaining over 98\% of its accuracy.
Under a comparable affected KV cache ratios, our method consistently outperforms existing KV cache management strategies 
like StreamingLLM~\cite{streaming-llm}, SnapKV~\cite{snapkv}, and Quest~\cite{quest} in terms of accuracy.

We summarize our key contributions as follows:

\begin{itemize}
    \item We show that LRMs frequently generate highly similar intermediate reasoning steps (\textit{redundant thinking}), suggesting that much of the KV cache content is similar. 
    \item We introduce ReasonCache, a novel KV cache management method that efficiently identifies and reuses similar KV cache blocks. 
    \item We comprehensively evaluate ReasonCache on a variety of tasks. The results show that ReasonCache achieves up to 89.2\% throughput gains while maintaining model accuracy.
\end{itemize}

\section{Background on LLM Inference}

The inference process of Large Language Models (LLMs) comprises two distinct stages: prefill and decoding~\cite{splitwise,distserve}, which also apply to LRMs.

\noindent \textbf{Compute-Bound Prefill Stage:}
During the prefill stage, the model processes the entire input sequence in parallel, 
performing computations such as matrix multiplications (GEMM) and attention calculations. 
The computational complexity of the attention mechanism scales quadratically with the sequence length,
 as it requires calculating interactions between all pairs of tokens~\cite{efficient-attention}.
%  \update{attention computation's equation is deleted}

%  The attention computation in the prefill stage can be formalized as follows. Let 
%  \(
%  X = (x_1, \ldots, x_n) \in \mathbb{R}^{n \times d}
%  \)
%  denote the input sequence of hidden states, where \(n\) is the sequence length and \(d\) is the hidden state dimension. For each attention head \(h\), the queries \(Q_h\), keys \(K_h\), and values \(V_h\) are computed as:
%      $Q_h = X W_q^h, K_h = X W_k^h, V_h = X W_v^h,$
%  where \(W_q^h, W_k^h, W_v^h \in \mathbb{R}^{d \times d_h}\) are the projection matrices for head \(h\), and \(d_h\) is the dimension of each head. The attention weights \(A_h\) are then calculated as:
% $ A_h = \mathrm{softmax}\left( \frac{Q_h (K_h)^\top}{\sqrt{d_h}} \right) $. 
%  Finally, the output of the attention head \(O_h\) is obtained by multiplying the attention weights with the values:
%      $O_h = A_h V_h.$

\noindent \textbf{Memory-Bound Decoding Stage:}
During the decoding stage, the computation of attention weights and output vectors involves matrix-vector 
operations with low arithmetic intensity while requiring access to the entire historical KV cache.
This creates a fundamental imbalance between the small amount of computation per step and the large memory required to store all KV cache.  %\update{deleted an example}
% The size of the KV cache scales with both sequence length and batch size. 
% For example, the KV cache footprint of a 32B-parameter Qwen model is substantial:
%  each token consumes about 1280 KB of memory\footnote{Calculated as 2 (key/value vectors) x 5120 (hidden dimensions) x 64 (number of layers) x 2  (size of FP16)},
% thus a single 10K-token request requires 12.2GB of memory for KV cache alone. The substantial memory usage severely limits the number of concurrent requests on GPUs.

\noindent \textbf{PagedAttention:} 
Driven by the pioneering work of vLLM~\cite{vllm} and SGLang~\cite{sglang}, PagedAttention has become the dominant technology for high performance LLM inference.
In PagedAttention, the KV cache is partitioned into fixed size blocks. 
This shifts the granularity of KV cache management from the entire sequence to individual blocks, significantly improving memory efficiency. 
Importantly, while blocks can be stored non-contiguously in memory, the order of tokens within each block remains fixed, preserving the sequential structure required for attention computation.

\section{Motivation}
In this section, we first show that redundant thinking is common in LRMs, then examine its impact on KV cache patterns. Our two key observations are as follows:

\noindent \textbf{Observation 1: \textit{Redundant thinking frequently occurs in the reasoning process of LRMs.}} 
During the reasoning process of LRMs, 
we observe the frequent emergence of highly similar or identical reasoning patterns, 
a phenomenon we term \textit{redundant thinking}. 
Figure \ref{fig1} provides empirical examples of redundant thinking across various LRMs. We summarize the patterns into three types:
(1) self-verification, where the model examines its previously generated content; 
(2) state revisitation, where the model repeatedly explores the same reasoning steps, often due to a failure to track its exploratory history;
 and (3) infinite self-loop, where the model fixes on two or more competing solutions, oscillating between them until the token budget is exhausted.

\begin{figure}[t]
    \centering
    \includegraphics[width=0.95\columnwidth]{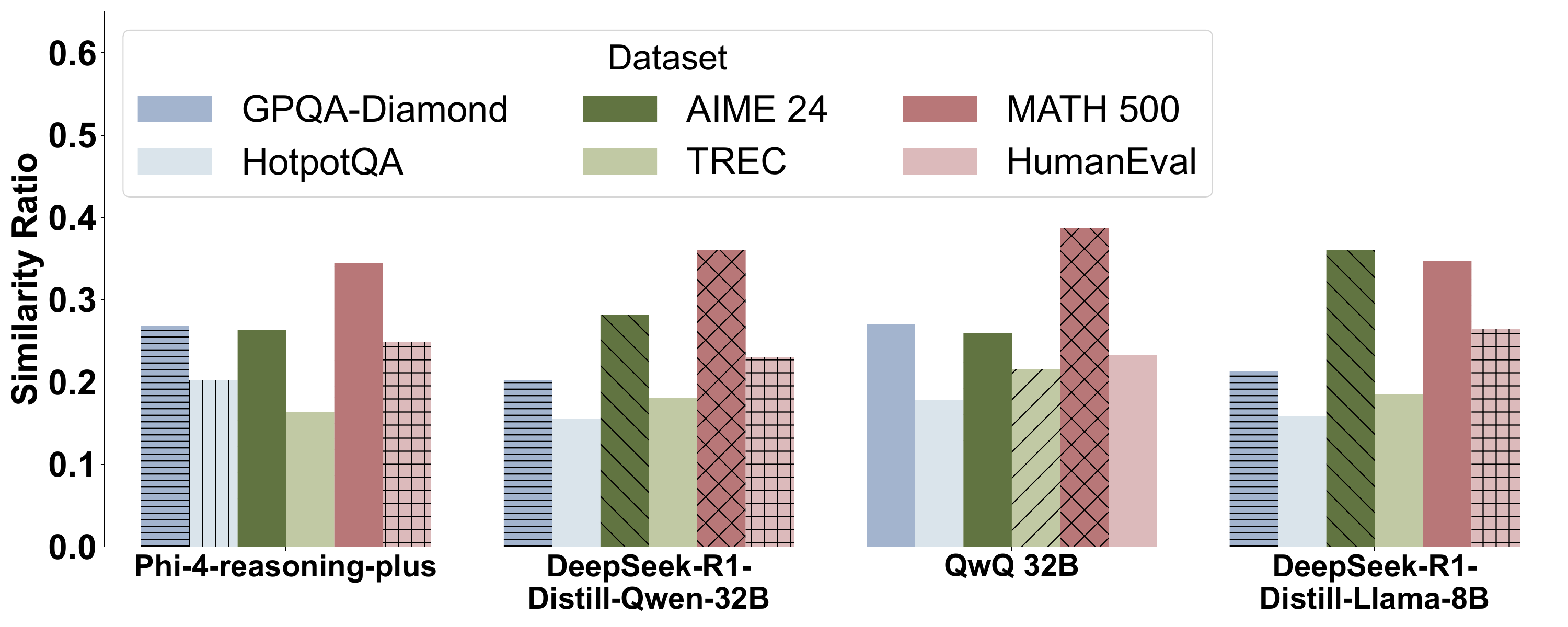}
    \caption{Redundant thinking across different reasoning models and datasets.}
    \label{fig2}
    \vspace{-4mm}
    \end{figure}

To quantitatively evaluate this phenomenon, we assess the redundancy in solutions generated by various models across multiple benchmarks. 
The reasoning process is typically organized into discrete steps, often separated by delimiters such as \texttt{\textbackslash n\textbackslash n}. 
Our methodology represents each reasoning step as a bag-of-words vector, then computes the cosine similarity between each step and all preceding steps. 
To measure overall redundancy, we introduce the \textbf{similarity ratio}, defined as the proportion of reasoning steps with a cosine similarity exceeding 0.8 to any prior step, 
indicating substantial lexical overlap.
As shown in Figure \ref{fig2}, our findings indicate that 15\% to 40\% of the generated content is redundant. 
This issue of redundant thinking appears to be consistent across different models and tasks, suggesting it is both model- and task-agnostic.

\begin{figure}[t] 
      \centering
      \begin{subfigure}[t]{0.49\linewidth}
          \includegraphics[width=\textwidth]{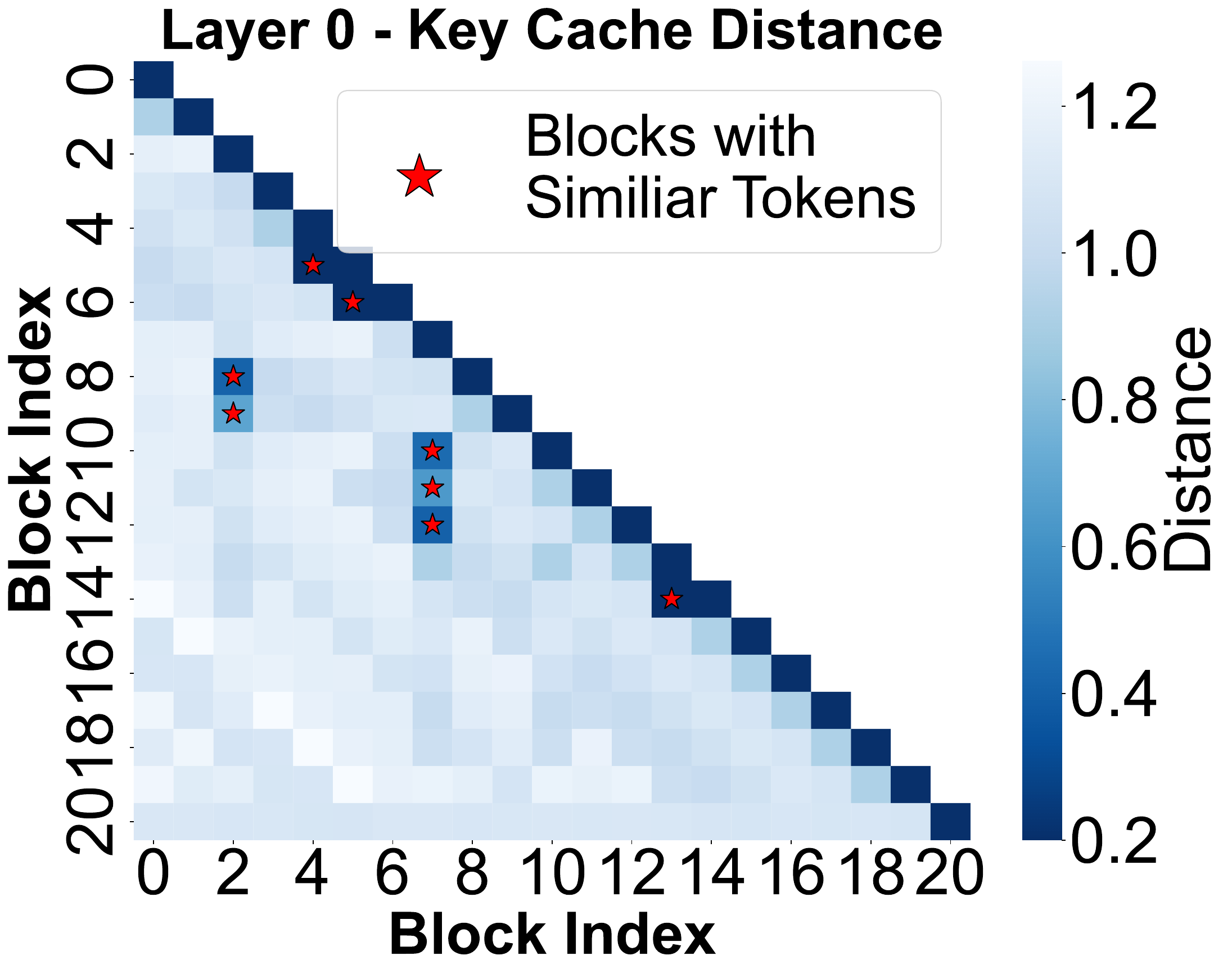}
      \end{subfigure}
      \hfill
      \begin{subfigure}[t]{0.49\linewidth}
          \includegraphics[width=\textwidth]{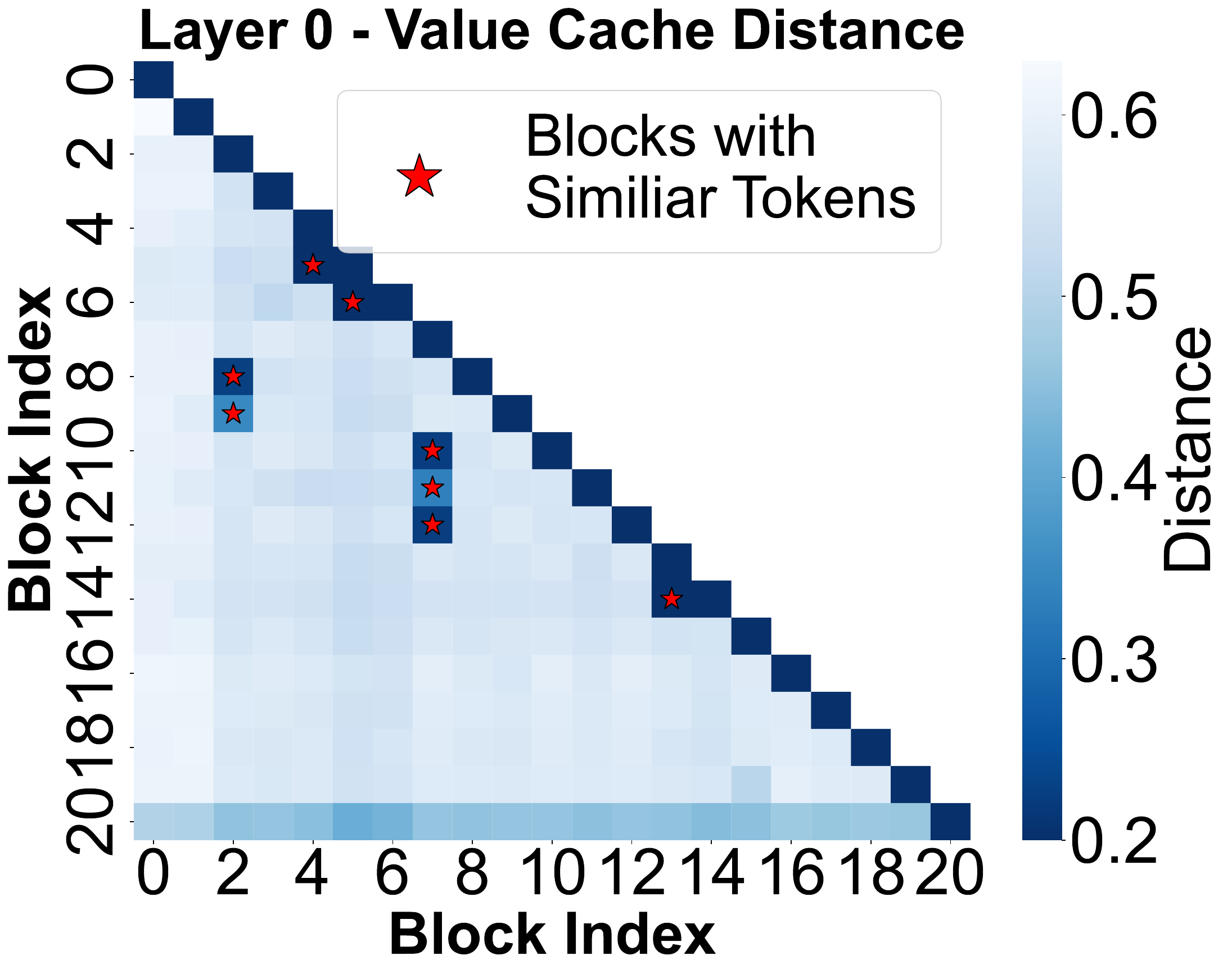}
      \end{subfigure}

      \begin{subfigure}[t]{0.49\linewidth}
          \includegraphics[width=\textwidth]{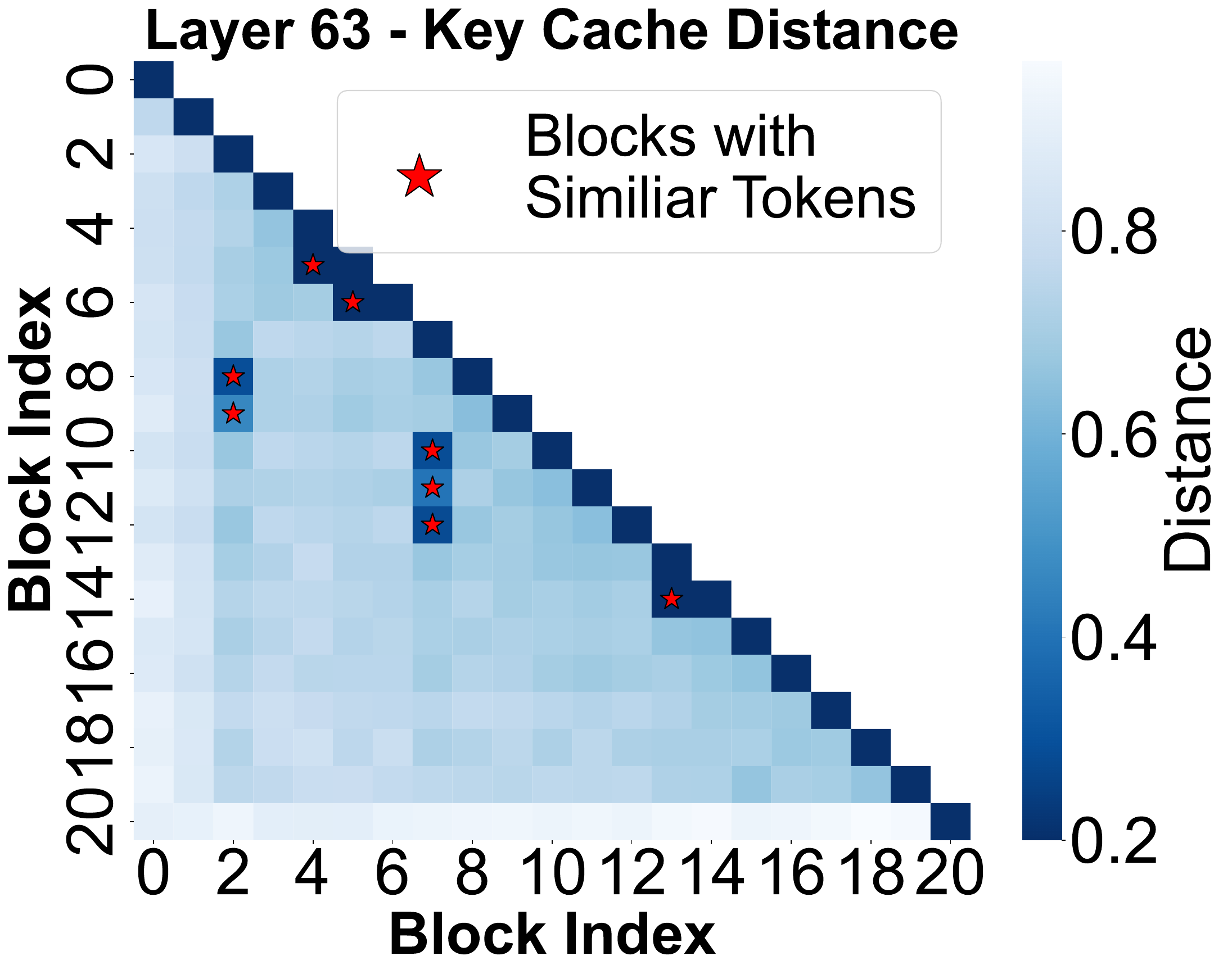}
      \end{subfigure}
      \hfill
      \begin{subfigure}[t]{0.49\linewidth}
          \includegraphics[width=\textwidth]{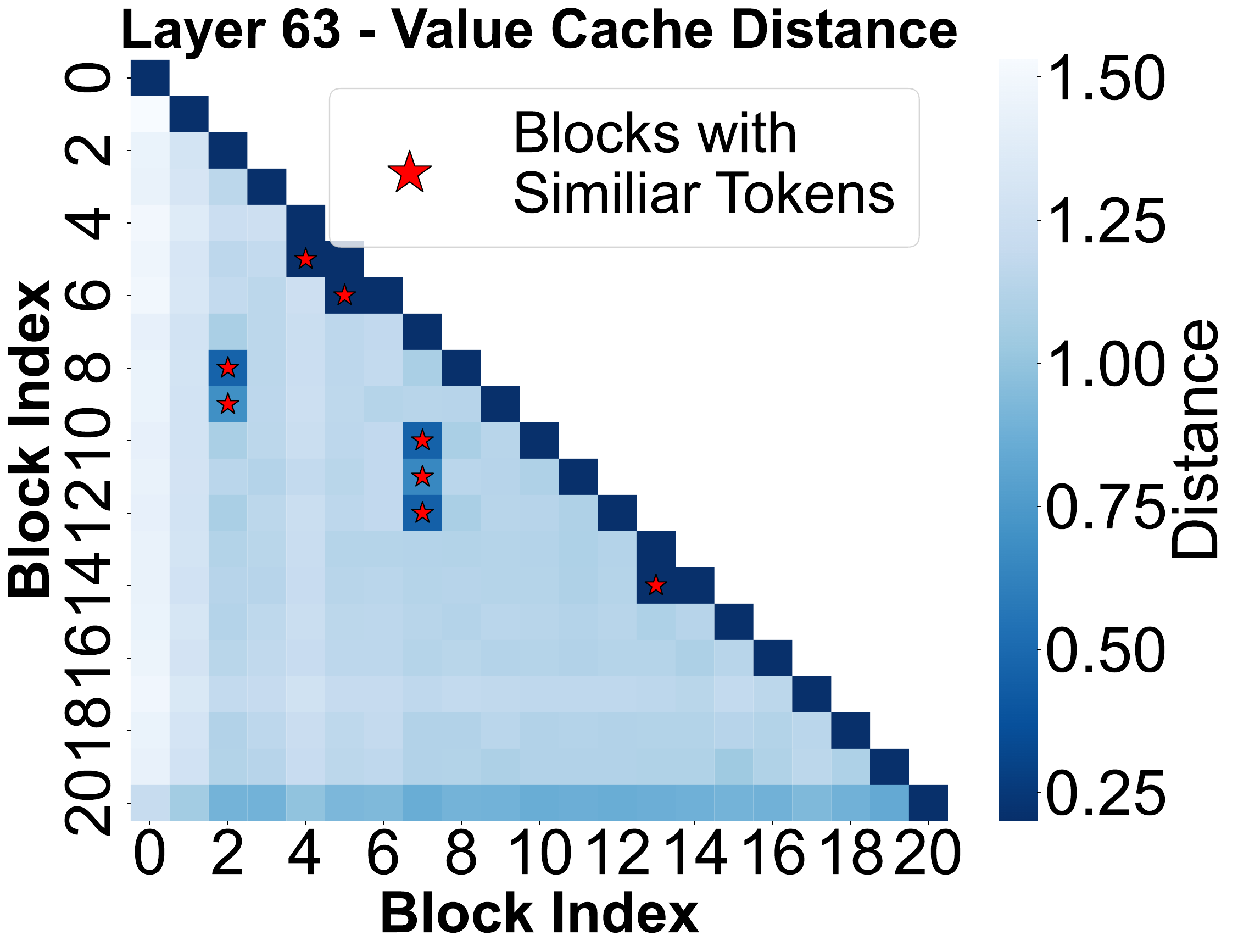}
      \end{subfigure}
      
      \caption{Block-wise Euclidean distance heatmaps for KV cache, 
      generated using QwQ-32B on the MATH 500 dataset. }
      \label{fig3}
      \vspace{-4mm}
    \end{figure}

    \noindent \textbf{Observation 2: \textit{Redundant thinking leads to similar KV cache patterns in LRMs.}}
    Inspired by prior studies on token similarity in attention mechanisms~\cite{d2o,ModelMerger}, 
    we hypothesize that the redundant thinking patterns in LRMs also manifest as similar KV cache states.
    To validate this, we analyze block-wise Euclidean distance between KV caches.
    Our findings reveal a clear pattern: as shown in Figure~\ref{fig3}, the heatmaps of key and value cache distances for layers 0 and 63 consistently show that lexically similar tokens (marked with red stars) have smaller Euclidean distances (deep bluer regions).
    Out of 21 blocks in total, 8 (approximately 38\%) are found to have a corresponding block that is both lexically similar and has a low Euclidean distance, 
    indicating their potential for reuse.
    % Further details, including experiment settings and results across different models and datasets, are provided in the Appendix~\ref{sec:supp_obs2}.	
    
    The above observations naturally motivate the idea of reusing KV cache blocks for similar reasoning steps to reduce memory overhead during inference.
    
    \begin{figure*}[h!]
        \centering
            \includegraphics[width=\textwidth]{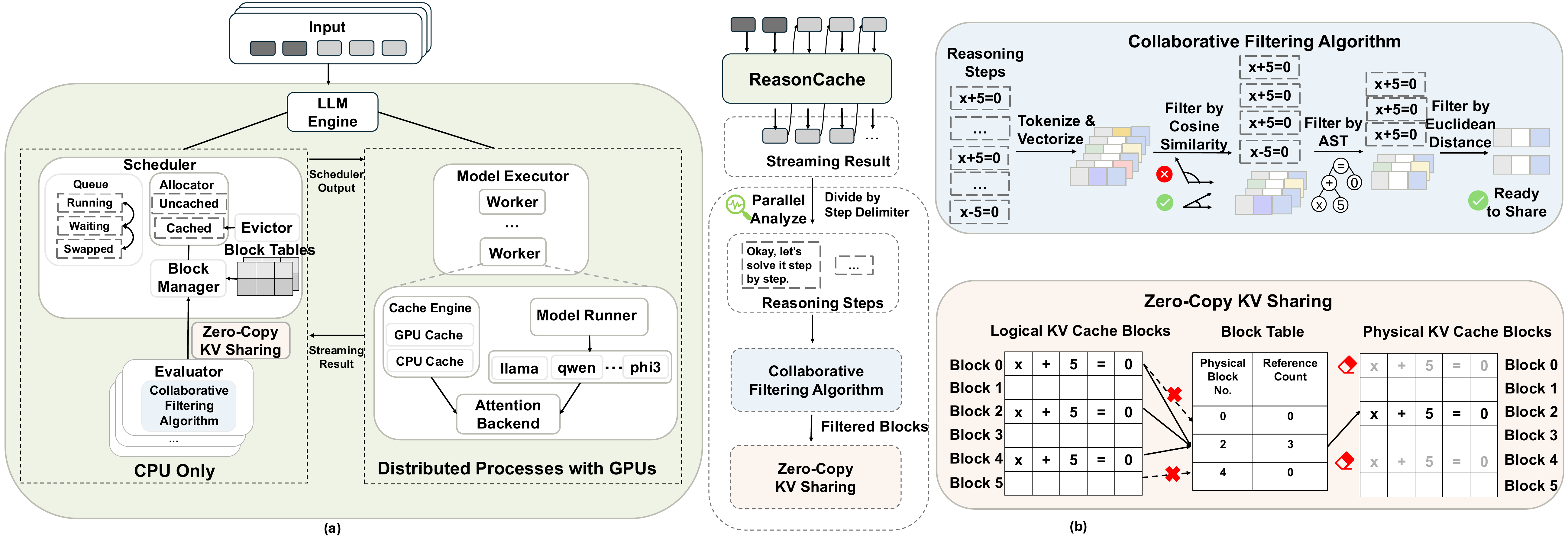}
            \caption{System architecture (a) and workflow (b) of ReasonCache. While the Language Reasoning Model (LRM) performs streaming token generation, ReasonCache executes Collaborative Filtering Algorithm and Zero-Copy KV Sharing in parallel. This non-blocking architecture reduces memory consumption and improves efficiency.}%\tx{\tx{For zero-copy KV sharing, maybe we could illustrate deduplication by removing redundant block-table entries and reclaiming the corresponding physical KV-cache blocks.}}}
            \label{fig4}
            \vspace{-4mm}
    \end{figure*}
    
    \section{ReasonCache}
    
    In this section, we introduce ReasonCache, a simple yet effective method designed to alleviate memory-bound constraints of LRM. 
    % thereby increasing batch size and throughput. 
    
    As illustrated in Figure \ref{fig4}, ReasonCache integrates two core components: 
    Collaborative Filtering Algorithm for efficiently and accurately identifying reusable KV cache blocks and Zero-Copy KV Sharing mechanism for seamless reuse. 
    The Collaborative Filtering Algorithm employs a two-stage process. It begins with a fast lexical and structural comparison of reasoning steps to identify potential candidates, 
    followed by a precise Euclidean distance calculation on their corresponding KV cache blocks to confirm reusability (Section~\ref{design:collaborative_filtering_algorithm}).
    To implement the reuse, the KV sharing mechanism enables zero-copy sharing by integrating with vLLM's memory manager. 
    It uses reference counting to map multiple logical blocks to a single physical copy, thereby avoiding costly data transfers (Section~\ref{design:zero-copy_kv_sharing}).
    Together, these components enable memory efficient decoding and accelerate LRM serving without compromising accuracy.
    We also theoretically prove that our design will preserve the accuracy (Section~\ref{design:theoretical_Foundation}).

    \subsection{Collaborative Filtering Algorithm}
    \label{design:collaborative_filtering_algorithm}
    To leverage redundancy within the KV cache,
    ReasonCache employs a two-stage filtering algorithm that efficiently and accurately identifies reusable blocks. 
    This entire process is designed as an asynchronous, lightweight filter offloaded to the CPU, ensuring it does not hinder GPU computation.
    The first stage performs coarse-grained filtering based on the textual content of reasoning steps. 
    It uses lexical and structural analysis to rapidly prune the vast number of preceding steps into a small set of promising candidates.
    The second stage then performs a fine-grained verification on this reduced set, operating directly on the KV cache blocks. 
    It calculates the Euclidean distance between their tensor representations to precisely quantify similarity, ensuring that only genuinely similar blocks are reused.
    \subsubsection{\textbf{Stage 1: Step-Level Similarity Filtering}}
    
    This stage implements a two-phase filtering process to identify candidate steps for KV cache reuse, 
    starting with a lexical pre-filter and then in a rigorous structural analysis.
    
    \noindent \textbf{Lexical Similarity Pre-filtering:} In the first phase, the lexical pre-filter assesses the similarity between the current step, $S_c$, and each candidate step, $S_k$, from all the preceding steps. 
    Specifically, each step is converted into a bag-of-words vector based on the frequency of its token IDs from the model's tokenizer. 
    For a vocabulary of size $|V|$, this process yields sparse vectors $\mathbf{v}_c, \mathbf{v}_k \in \mathbb{R}^{|V|}$, where each dimension counts the occurrences of a corresponding token ID.
    The similarity score is then calculated as the cosine similarity between these vectors, adjusted by a length penalty:
    {
    \small
    \label{eq:cosine_similarity}
    $$\text{sim}(S_c, S_k) = \frac{\mathbf{v}_c \cdot \mathbf{v}_k}{\|\mathbf{v}_c\| \|\mathbf{v}_k\|} \times \frac{\min(|S_c|, |S_k|)}{\max(|S_c|, |S_k|)}$$
    }
    % \begin{equation}
    % \label{eq:cosine_similarity}
    % \text{sim}(S_c, S_k) = \frac{\mathbf{v}_c \cdot \mathbf{v}_k}{\|\mathbf{v}_c\| \|\mathbf{v}_k\|} \times \frac{\min(|S_c|, |S_k|)}{\max(|S_c|, |S_k|)}
    % \end{equation}
    where $|S|$ represents the character length of the text in a step. This length penalty favors candidates of similar length to the current step, mitigating bias towards shorter sequences which might otherwise achieve high cosine similarity with fewer token overlaps.
    
    A candidate step $S_k$ is retained for further analysis if its similarity score exceeds a dynamic threshold $\tau$, i.e., $\text{sim}(S_c, S_k) > \tau$. 
    The dynamic threshold $\tau$ is adapted based on the similarity scores between the current step $S_c$ and all preceding steps $\{S_1, \dots, S_{c-1}\}$. 
    Specifically, let $\mathcal{H}c = \{\text{sim}(S_c, S_i) \mid i=1, \dots, c-1\}$ be the set of historical similarity scores. 
    The threshold $\tau$ is then defined as:
    % \begin{equation}
    {
    \label{eq:dynamic_threshold}
    $$\tau = \tau_{\text{strict}} - (\tau_{\text{strict}} - \tau_{\text{soft}}) \times \mu_{\mathcal{H}c}$$}
    % \end{equation}
    
    where $\mu_{\mathcal{H}c}$ represent the mean of the similarity scores in $\mathcal{H}_c$. 
    $\tau_{\text{strict}}$ is the highest, strictest threshold, while $\tau_{\text{soft}}$ is the lowest, most lenient threshold.
    
    The purpose of this dynamic threshold is to adaptively adjust the reuse policy based on the model's current reasoning pattern. 
    When the model enters a highly repetitive state (e.g., repeatedly verifying the same formula), the similarity scores with preceding steps are consistently high, which elevates the mean, $\mu_{\mathcal{H}c}$. 
    Consequently, the threshold $\tau$ is lowered towards $\tau_{\text{soft}}$, making it more lenient. 
    Conversely, when the model explores a novel reasoning path, similarity scores are generally low. This depresses $\mu_{\mathcal{H}c}$ and raises $\tau$ towards its strictest value, $\tau_{\text{strict}}$. 
    This strict criterion ensures that only exceptionally similar steps are reused.
    
    \noindent \textbf{Structural Equivalence Verification:} Candidates that successfully pass the lexical filter advance to the structural analysis phase.
    It is important to note that this verification is selectively applied based on the content type.
    It targets steps expressed in formal languages (e.g., code, math) where structural precision is critical to prevent semantic errors (e.g., distinguishing \texttt{2x+5=10} from \texttt{2x-5=10}).
    For unstructured content like open-domain dialogue, where ASTs are undefined, this step is bypassed.

    For formal content, steps are parsed into their corresponding Abstract Syntax Trees (ASTs)~\cite{AST}, denoted as $T = \mathcal{P}(S)$. 
    An AST is a labeled tree $T=(V, E, L)$, where the nodes $V$ represent components of the expression, the edges $E$ define operator-operand relationships, and $L$ is a labeling function that assigns a symbol (e.g., operator, variable, constant) to each node.
    
    To compare the structural properties of the current step $S_c$ and a candidate step $S_k$, their respective ASTs, $T_c = \mathcal{P}(S_c)$ and $T_k = \mathcal{P}(S_k)$, are generated. The structural divergence between these trees is quantified using the Tree Edit Distance (TED)~\cite{TED}, denoted as $\mathcal{D}(T_c, T_k)$. TED measures the minimum cost required to transform $T_c$ into $T_k$ through a sequence of three fundamental node operations: insertion, deletion, and substitution. Formally, let $\gamma(u \to v)$ be the cost of substituting node $u$ with $v$, $\gamma(u \to \lambda)$ be the cost of deleting $u$, and $\gamma(\lambda \to v)$ be the cost of inserting $v$. The TED is then the minimum cumulative cost over all valid edit mappings between the nodes of $T_c$ and $T_k$.
        {
        \small
        $$\mathcal{D}(T_c, T_k) = \min_{M} \Biggl( \quad \sum_{\mathclap{(u,v) \in M}} \gamma(u \to v)  + \sum_{\mathclap{u \in V_c \setminus \text{dom}(M)}} \gamma(u \to \lambda) + \sum_{\mathclap{v \in V_k \setminus \text{ran}(M)}} \gamma(\lambda \to v) \Biggr)$$
        }
    where $V_c$ and $V_k$ are the node sets of $T_c$ and $T_k$ respectively, 
    and the mapping $M$ must preserve the ancestry relationships between nodes. 
    A candidate is discarded if $\mathcal{D}(T_c, T_k) > 0$, as this indicates a mismatch in their underlying structure for formal language.
    
    \subsubsection{\textbf{Stage 2: Block-Level Distance Filtering}}
    
    For steps flagged as similar in Stage 1, this stage computes the Euclidean distance between their corresponding KV cache blocks. 
    This process comprises two key components: an optimized distance computation method and an adaptive thresholding policy.
    
    \noindent \textbf{Distance Computation:}
    Let the KV cache blocks for the current step $S_c$ and a candidate step $S_k$ be denoted as $\mathcal{B}_c$ and $\mathcal{B}_k$. 
    The computation aggregates normalized layer-wise distances across all $N$ layers of the model. 
    For each layer $i \in \{1, \dots, N\}$, the Euclidean distance between the key and value tensors of the two blocks is calculated:
    {\small
    $$\Delta_{K,i} = \|\mathbf{K}_i[\mathcal{B}_c] - \mathbf{K}_i[\mathcal{B}_k]\|_2 ,\quad \Delta_{V,i} = \|\mathbf{V}_i[\mathcal{B}_c] - \mathbf{V}_i[\mathcal{B}_k]\|_2$$
    }
    % \begin{align}
    %     \Delta_{K,i} &= \|\mathbf{K}_i[\mathcal{B}_c] - \mathbf{K}_i[\mathcal{B}_k]\|_2 \\
    %     \Delta_{V,i} &= \|\mathbf{V}_i[\mathcal{B}_c] - \mathbf{V}_i[\mathcal{B}_k]\|_2
    % \end{align}
    where $\mathbf{K}_i[\mathcal{B}]$ and $\mathbf{V}_i[\mathcal{B}]$ extracts the key and value tensor for block $\mathcal{B}$ from the layer $i$ cache, respectively, and $\|\cdot\|_2$ is the L2 norm of the flattened tensor.
    
    The final normalized distance $\bar{d}$ is the average of these layer-wise distances. Each layer's distance is computed as the sum of its key and value distances, normalized by the block's dimensions:
        {\small        
        \label{eq:block_dist}
        $$\bar{d}(\mathcal{B}_c, \mathcal{B}_k) = \frac{1}{N} \sum_{i=1}^{N} \frac{\Delta_{K,i} + \Delta_{V,i}}{2dh}$$
        }    
    where $d$ is the number of tokens in the blocks (i.e., block size), and $h$ is the number of KV heads. 
    
    To amortize the cost of distance computation and maximize reuse, we can expand the squared Euclidean distance:
    % \begin{equation}
{   \small 
    \label{eq:squared_euclidean_reuse}
    $$\|\mathbf{x}_c - \mathbf{x}_k\|^2 = \|\mathbf{x}_c\|^2 + \|\mathbf{x}_k\|^2 - 2\, \mathbf{x}_c^\top \mathbf{x}_k.$$}
    % \end{equation}
    Here, $\mathbf{x}_c$ and $\mathbf{x}_k$ represent the flattened key or value tensors from the blocks being compared 
    (e.g., $\mathbf{x}_c = \mathbf{K}_i[\mathcal{B}_c]$ and $\mathbf{x}_k = \mathbf{K}_i[\mathcal{B}_k]$). 
    The squared norms $\|\mathbf{x}_c\|^2$ and $\|\mathbf{x}_k\|^2$ can be computed once for each block and cached for long-term reuse. 
    We employ lazy computation: when a block first becomes a reuse candidate, we perform a single reduction operation to compute its tensor norms, storing these values in an LRU cache attached to the block metadata. 
    Subsequent distance measurements reuse the cached norms and only require computing the dot product $\mathbf{x}_c^\top \mathbf{x}_k$.
    
    \noindent \textbf{Percentile-based Adaptive Thresholding:} To assess the reusability of a KV cache block, we adopt Percentile-based Adaptive Thresholding (PAT). 
    This method sets a dynamic threshold based on the quantile of the current distance distribution, making it robust across different models and tasks without a fixed value. 
    The system operates in two phases: (1) it collects distance measurements to establish a baseline distribution (KV cache reuse is disabled during this warm-up period).
    (2) it uses the dynamically computed \(p\)th percentile as the threshold for all sharing decisions. 
    
    The choice of PAT is motivated by the model-dependent nature of KV cache similarity metrics. 
    The distance between KV cache entries can vary with model architecture, training data, and scale. 
    Consequently, a fixed distance threshold for determining reusability is suboptimal, as a value indicating high similarity in one model may represent only moderate similarity in another. 
    ReasonCache addresses this challenge by setting the dynamic threshold at a specific percentile, \(p\), of this observed distribution for all subsequent sharing decisions. 
    This ensures that the criterion for reuse is always relative to the specific model's behavior, making the approach robust and model-agnostic.

\noindent \textbf{Complexity Analysis:}
We analyze the time complexity of ReasonCache.
Let $n$ denote the total token length and $m$ denote the total number of reasoning steps ($m \approx n/|S|$, where $|S|$ is the average step length).
For each step, the algorithm performs candidate filtering and similarity verification.
The total computational cost is $O(m \cdot k \cdot (|S| + N \cdot d \cdot h))$, where $k$ denotes the number of candidates after filtering ($k \ll m$), $N$ is the number of layers, $d$ is the block size, and $h$ is the number of attention heads.
Since $N, d$, and $h$ are architecture-specific constants, the complexity depends primarily on $m$ and $k$.
The effective complexity scales as $O(n \log n)$ in practice.
This is because the number of reasoning steps $m$ grows linearly with $n$, while the number of candidates $k$ typically grows logarithmically ($k \propto \log n$) due to the high selectivity of the two-stage filtering process, which efficiently prunes the search space.
Furthermore, ReasonCache is implemented in a \textit{non-blocking and asynchronous} manner, running Collaborative Filtering Algorithm on the CPU in parallel with the GPU inference.
This design effectively hides the overhead, ensuring that the additional computation does not block the critical path of token generation.

    \subsection{Zero-Copy KV Sharing}
    \label{design:zero-copy_kv_sharing}
    Figure \ref{fig4} illustrates our KV cache sharing mechanism, which is designed to integrate seamlessly with PagedAttention by operating at the same block-level granularity. 
    A physical block is marked as sealed and becomes immutable once fully filled with tokens, making it eligible for sharing.
    The evaluator identifies sharing opportunities using Collaborative Filtering Algorithm. 
    Instead of performing costly data copies, sharing is achieved by incrementing the block's reference count and updating the block table accordingly.
    A physical block is returned to the free pool by the evictor only when its reference count drops to zero.
    This sharing process is integrated into the standard decoding loop: during each decoding step, the scheduler batches requests, 
    organizes their block table information, and forwards them to the model executor for inference. 
    
    A key advantage of this approach is its ability to leverage vLLM's decoupled CPU-GPU architecture. 
    The GPU model executor can immediately act on sharing decisions through simple updates to the CPU-managed block table, 
    thereby reusing cache blocks without interrupting the autoregressive generation process or consuming additional GPU memory bandwidth.
    
    \subsection{Theoretical Foundation}
    \label{design:theoretical_Foundation}
    The central premise of ReasonCache is that reusing KV cache blocks that are similar, but not identical, can yield substantial efficiency gains. 
    This premise, however, raises a critical question: \textbf{Does reusing similar but not identical KV cache blocks adversely affect model accuracy?}
    To formally prove the validity of our approach, we present Theorem 1.
    
    \noindent \textbf{Theorem 1.} \textit{Let $q_t$ be a query vector at step $t$, and let $K_t = [k_1, k_2, \ldots, k_t]$ and $V_t = [v_1, v_2, \ldots, v_t]$ 
    denote the key and value caches, respectively. 
    Suppose $k_j$ is replaced by   $k_j'$ such that $\|k_j - k_j'\|_2 < \varepsilon$, and $v_j$ is replaced by $v_j'$ such that $\|v_j - v_j'\|_2 < \delta$. Then:}
    
    The change in the attention score for position $j$ satisfies
        \[
        |q_t k_j^\top - q_t {k_j'}^\top| = |q_t(k_j^\top - {k_j'}^\top)|  \leq \|q_t\|_2 \cdot \|k_j - k_j'\|_2 < \|q_t\|_2 \varepsilon.
        \]
        The first inequality follows the Cauchy-Schwarz inequality. Consequently, the resulting attention weight $A^t = \mathrm{softmax}\left(\frac{q_t K_t^\top}{\sqrt{d}}\right)$ is only minimally perturbed, as the softmax function is Lipschitz continuous~\cite{lipschitz,function-lipschitzness}.
        
     The change in the output $o_t = \sum_{j=1}^t A_j^t v_j$ due to replacing $v_j$ with $v_j'$ is bounded by
        \[
        \|A_j^t v_j - A_j^t v_j'\|_2 = \|A_j^t (v_j -  v_j')\|_2=  A_j^t \|v_j - v_j'\|_2 < A_j^t \delta,
        \]
        and the total perturbation in $o_t$ is at most $\delta$ since $0<A_j^t<1$.
    
    Therefore, substituting a key or value vector with a sufficiently close alternative induces only minor changes in the attention output, thereby justifying the validity of KV cache sharing. 
    Rigorous proofs are provided in the Appendix~\ref{sec:proof}, and empirical results are presented in the evaluation section.

    \begin{table*}[h!]
        \centering
        \footnotesize
        \setlength{\tabcolsep}{0.6mm}
        \caption{Accuracy and throughput (tokens/s) evaluation. Bold values represent the throughput improvement relative to Dense.} 
        \resizebox{\textwidth}{!}{%
        \begin{tabular}{cccccccccccccccccc}
            \toprule
            & & \multicolumn{2}{c}{\textit{Code}} & \multicolumn{2}{c}{\textit{Multi-doc QA}} & \multicolumn{2}{c}{\textit{Summarization}} & \multicolumn{6}{c}{\textit{Math}} & \multicolumn{2}{c}{\textit{Science}} \\
            \cmidrule(lr){3-4} \cmidrule(lr){5-6} \cmidrule(lr){7-8} \cmidrule(lr){9-14}  \cmidrule(lr){15-16}
            \multirow{1}{*}{\makecell{Model}} & \multirow{1}{*}{\makecell{Method}} & \multicolumn{2}{c}{\makecell{HumanEval}} & \multicolumn{2}{c}{\makecell{HotpotQA}} & \multicolumn{2}{c}{\makecell{TREC}} & \multicolumn{2}{c}{\makecell{AIME 2024}} & \multicolumn{2}{c}{\makecell{AIME 2025}}& \multicolumn{2}{c}{\makecell{MATH 500}} & \multicolumn{2}{c}{\makecell{GPQA Diamond}} &\makecell{Avg\\Acc} &\makecell{Avg \\throughput}\\
            % \cmidrule(lr){3-4} \cmidrule(lr){5-6} \cmidrule(lr){7-8} \cmidrule(lr){9-10} \cmidrule(lr){11-12} \cmidrule(lr){13-14} \cmidrule(lr){15-16}
            & & \makecell{Acc} & \makecell{throughput} & \makecell{Acc} & \makecell{throughput} & \makecell{Acc} & \makecell{throughput} & \makecell{Acc} & \makecell{throughput} & \makecell{Acc} & \makecell{throughput} & \makecell{Acc} & \makecell{throughput} & \makecell{Acc} & \makecell{throughput} \\
            \midrule
            \multirow{4}{*}{\makecell{DeepSeek R1\\ Distill\\Qwen 32B}}
            & \makecell{ReasonCache} & \makecell{0.969} & \makecell{\textbf{(+62.36\%)}\\246.01} & \makecell{0.36} & \makecell{\textbf{(+42.18\%)}\\542.43} & \makecell{0.61} & \makecell{\textbf{(+58.87\%)}\\416.84} & \makecell{0.7} & \makecell{\textbf{(+72.38\%)}\\123.4} & \makecell{0.5} & \makecell{\textbf{(+43.1\%)}\\82.34} & \makecell{0.918} & \makecell{\textbf{(+89.2\%)}\\194.5} & \makecell{0.607} & \makecell{\textbf{(+56.6\%)}\\176.21} & \makecell{0.665} & \makecell{\textbf{(+56.33\%)}\\254.53}\\
            & \makecell{Random} & \makecell{0.622} & \makecell{\textbf{(+63.09\%)}\\247.11} & \makecell{0.18} & \makecell{\textbf{(+40.98\%)}\\537.85} & \makecell{0.28} & \makecell{\textbf{(+54.89\%)}\\406.40} & \makecell{0.367} & \makecell{\textbf{(+68.73\%)}\\120.8} & \makecell{0.2} & \makecell{\textbf{(+42.2\%)}\\81.81} & \makecell{0.6} & \makecell{\textbf{(+85.9\%)}\\191.2} & \makecell{0.323} & \makecell{\textbf{(+53.8\%)}\\172.91}  & \makecell{0.367} & \makecell{\textbf{(+54.25\%)}\\251.15}\\
            & \makecell{Dense} & \makecell{0.982} & \makecell{151.52} & \makecell{0.375} & \makecell{381.51} & \makecell{0.62} & \makecell{262.38} & \makecell{0.733} & \makecell{71.59} & \makecell{0.533} & \makecell{57.55} & \makecell{0.927} & \makecell{102.82} & \makecell{0.631} & \makecell{112.40}  & \makecell{0.686} & \makecell{162.82}\\
           \midrule
            \multirow{4}{*}{QwQ 32B} 
            & \makecell{ReasonCache} & \makecell{0.972} & \makecell{\textbf{(+72.31\%)}\\193.52} & \makecell{0.41} & \makecell{\textbf{(+42.34\%)}\\423.23} & \makecell{0.652} & \makecell{\textbf{(+61.23\%)}\\287.59} & \makecell{0.733} & \makecell{\textbf{(+78.27\%)}\\85.3} & \makecell{0.633} & \makecell{\textbf{(+68.9\%)}\\82.3} & \makecell{0.872} & \makecell{\textbf{(+81.6\%)}\\152.31} & \makecell{0.581} & \makecell{\textbf{(+59.0\%)}\\125.67} & \makecell{0.693} & \makecell{\textbf{(+59.29\%)}\\192.85} \\
            & \makecell{Random} & \makecell{0.598} & \makecell{\textbf{(+70.67\%)}\\191.68} & \makecell{0.22} & \makecell{\textbf{(+41.89\%)}\\421.89} & \makecell{0.333} & \makecell{\textbf{(+60.89\%)}\\286.98} & \makecell{0.3} & \makecell{\textbf{(+79.1\%)}\\85.7} & \makecell{0.233} & \makecell{\textbf{(+72.6\%)}\\84.1} & \makecell{0.588} & \makecell{\textbf{(+82.7\%)}\\153.23} & \makecell{0.278} & \makecell{\textbf{(+55.9\%)}\\123.24} & \makecell{0.364} & \makecell{\textbf{(+58.92\%)}\\192.40} \\
            & \makecell{Dense} & \makecell{0.968} & \makecell{112.31} & \makecell{0.45} & \makecell{297.34} & \makecell{0.648} & \makecell{178.37} & \makecell{0.766} & \makecell{47.85} & \makecell{0.633} & \makecell{48.72} & \makecell{0.894} & \makecell{83.85} & \makecell{0.611} & \makecell{79.06} & \makecell{0.71} & \makecell{121.07} \\
            \midrule
            \multirow{4}{*}{\makecell{Phi 4\\ reasoning\\ plus}}
            & \makecell{ReasonCache} & \makecell{0.945} & \makecell{\textbf{(+51.23\%)}\\338.33} & \makecell{0.4} & \makecell{\textbf{(+32.45\%)}\\695.88} & \makecell{0.61} & \makecell{\textbf{(+42.13\%)}\\556.84} & \makecell{0.8} & \makecell{\textbf{(+45.85\%)}\\245.21} & \makecell{0.633} & \makecell{\textbf{(+44.7\%)}\\243.21} & \makecell{0.903} & \makecell{\textbf{(+65.5\%)}\\384.21} & \makecell{0.508} & \makecell{\textbf{(+39.0\%)}\\243.22} & \makecell{0.686} & \makecell{\textbf{(+43.3\%)}\\386.7} \\
            & \makecell{Random} & \makecell{0.482} & \makecell{\textbf{(+50.97\%)}\\337.75} & \makecell{0.21} & \makecell{\textbf{(+33.21\%)}\\699.87} & \makecell{0.315} & \makecell{\textbf{(+40.89\%)}\\552.08} & \makecell{0.333} & \makecell{\textbf{(+40.77\%)}\\243.2} & \makecell{0.133} & \makecell{\textbf{(+42.4\%)}\\239.41} & \makecell{0.492} & \makecell{\textbf{(+64.2\%)}\\381.23} & \makecell{0.203} & \makecell{\textbf{(+38.1\%)}\\241.81}  & \makecell{0.324} & \makecell{\textbf{(+42.69\%)}\\385.05}\\
            & \makecell{Dense} & \makecell{0.933} & \makecell{223.72} & \makecell{0.43} & \makecell{525.39} & \makecell{0.635} & \makecell{391.85} & \makecell{0.8} & \makecell{172.77} & \makecell{0.677} & \makecell{168.12} & \makecell{0.893} & \makecell{232.14} & \makecell{0.520} & \makecell{175.04}& \makecell{0.698} & \makecell{269.86}  \\
            \bottomrule
        \end{tabular}%
        }
        \label{table1}
        \vspace{-2mm}
      \end{table*}

    \section{Evaluation}
    \subsection{Settings}
    
    \noindent \textbf{Benchmarks:} 
    We evaluate ReasonCache on a diverse suite of benchmarks targeting distinct reasoning capabilities. 
    For coding, we use \textbf{HumanEval}~\cite{humanEval}, comprising 164 programming problems with unit tests to assess functional correctness. 
    For long-context understanding, we employ \textbf{HotpotQA}~\cite{hotpotqa} and \textbf{TREC}~\cite{trec} from LongBench~\cite{longbench} (200 samples each), which test multi-document reasoning and fine-grained classification, respectively. 
    Mathematical reasoning is evaluated on \textbf{AIME}~\cite{aime} and \textbf{MATH-500}~\cite{math500}, covering advanced algebra, geometry, and number theory. 
    Finally, \textbf{GPQA Diamond}~\cite{gpqa} assesses scientific reasoning with 198 graduate-level questions across biology, physics, and chemistry.

    \noindent \textbf{Models:}
    We employ three models with diverse sizes and architectures: DeepSeek-R1-Distill-Qwen-32B, QwQ-32B, and Phi-4-reasoning-plus to ensure a comprehensive evaluation.
    
    \noindent \textbf{Baselines:}
    We compare ReasonCache against a fully cached baseline (Dense) and other methods, including StreamingLLM~\cite{streaming-llm}, Quest~\cite{quest} and SnapKV~\cite{snapkv}.
    
     \noindent \textbf{Parameters:}
    % In our experimental setup, we utilize two thresholds for the step-level similarity score: 
    % $\tau_{\text{strict}} = 0.9$, $\tau_{\text{soft}} = 0.7$.
    %  For the block-level selection, we employ Percentile-based Adaptive Thresholding (PAT). 
    %  We set the PAT percentile to 80, meaning we retain 80\% of the candidate blocks with the lowest Euclidean distances for sharing.
     We use step-level similarity thresholds of $\tau_{\text{strict}} = 0.9$ and $\tau_{\text{soft}} = 0.7$. 
     For block-level selection, we employ Percentile-based Adaptive Thresholding (PAT), 
     retaining the 80\% of candidate blocks with the lowest Euclidean distances for sharing.
     
     \noindent \textbf{Metrics:}
     We evaluate the performance of different methods using three key metrics: Accuracy, Throughput, and Average Memory Saving Ratio (AMSR). 
    %  Accuracy measures the correctness of the model-generated results. 
     Throughput, measured in tokens per second (tokens/s), is a critical indicator of QoS in AI inference systems. 
     It directly reflects the system's processing capacity under high concurrency, where higher throughput translates to a greater ability to serve concurrent users.
    Average Memory Saving Ratio (AMSR) quantifies the memory efficiency of cache management strategies.
    For sharing, eviction and merging-based methods, it represents the physical reduction in KV cache memory usage.
    For selective loading methods like Quest~\cite{quest}, which retain the full KV cache, AMSR measures the fraction of KV cache blocks that are skipped and thus not loaded during the attention computation.

     \noindent \textbf{Testing Environment:}
    To simulate a realistic AI inference system under high load, 
    all benchmark tasks are submitted as concurrent requests to a vLLM-based engine,
     which dynamically schedules and batches them. 
     Experiments are conducted on NVIDIA A800 (80GB) GPUs with Ubuntu 22.04.3 and vLLM version 0.8.2. 
     All reported results are averaged over 5 independent runs.

      \subsection{Main Results}
      We compare ReasonCache against two baseline settings: Dense Attention, which reserves the full KV cache without sharing, and Random Sharing, which reuses KV cache blocks arbitrarily.
For a fair comparison, we align the Average Memory Saving Ratio (AMSR) of Random Sharing with that of ReasonCache.
As shown in Table \ref{table1}, ReasonCache delivers significant throughput gains across all evaluated models and benchmarks.
For example, it achieves average throughput improvements of 59.29\% for QwQ-32B.
Notably, these significant performance enhancements are realized with minimal impact on model accuracy.
For instance, Phi-4-reasoning-plus achieves a 43.3\% throughput boost while retaining 98.3\% of the baseline accuracy of Dense Attention.

\begin{table*}[h!]
    \centering
  %   \small
  %   \setlength{\tabcolsep}{0.8mm}
    \caption{Evaluation of different KV cache management strategies. Average Memory Saving Ratio (AMSR) quantifies the memory efficiency of a strategy. For each benchmark, the highest and second-highest accuracies are highlighted in bold and underlined, respectively.} 
    \resizebox{\textwidth}{!}{%
    \begin{tabular}{cccccccccccccccccc}
        \toprule
        & & \multicolumn{2}{c}{\textit{Code}} & \multicolumn{2}{c}{\textit{Multi-doc QA}} & \multicolumn{2}{c}{\textit{Summarization}} & \multicolumn{6}{c}{\textit{Math}} & \multicolumn{2}{c}{\textit{Science}} \\
        \cmidrule(lr){3-4} \cmidrule(lr){5-6} \cmidrule(lr){7-8} \cmidrule(lr){9-14}  \cmidrule(lr){15-16}
        \multirow{1}{*}{\makecell{Model}} & \multirow{1}{*}{\makecell{Method}} & \multicolumn{2}{c}{\makecell{HumanEval}} & \multicolumn{2}{c}{\makecell{HotpotQA}} & \multicolumn{2}{c}{\makecell{TREC}} & \multicolumn{2}{c}{\makecell{AIME 2024}} & \multicolumn{2}{c}{\makecell{AIME 2025}}& \multicolumn{2}{c}{\makecell{MATH 500}} & \multicolumn{2}{c}{\makecell{GPQA Diamond}} & {\makecell{Avg \\Acc}} & {\makecell{AMSR}} \\
      %   & & \multicolumn{2}{c}{\textit{(Code)}} & \multicolumn{2}{c}{\textit{(Multi-doc QA)}} & \multicolumn{2}{c}{\textit{(QA)}} & \multicolumn{2}{c}{\textit{(Math)}} & \multicolumn{2}{c}{\textit{(Math)}} & \multicolumn{2}{c}{\textit{(Math)}} & \multicolumn{2}{c}{\textit{(Science)}} \\
      %   \cmidrule(lr){3-4} \cmidrule(lr){5-6} \cmidrule(lr){7-8} \cmidrule(lr){9-10} \cmidrule(lr){11-12} \cmidrule(lr){13-14} \cmidrule(lr){15-16}
        & & \makecell{Acc} & \makecell{AMSR} & \makecell{Acc} & \makecell{AMSR} & \makecell{Acc} & \makecell{AMSR} & \makecell{Acc} & \makecell{AMSR} & \makecell{Acc} & \makecell{AMSR} & \makecell{Acc} & \makecell{AMSR} & \makecell{Acc} & \makecell{AMSR} \\
        \midrule
        \multirow{5}{*}{\makecell{DeepSeek R1\\ Distill\\Qwen 32B}}
        & \makecell{StreamingLLM} & \makecell{0.884} & \makecell{20.1\%} & \makecell{0.235} & \makecell{15.8\%} & \makecell{0.47} & \makecell{15.5\%} & \makecell{0.567} & \makecell{22.34\%} & \makecell{0.333} & \makecell{20.7\%} & \makecell{0.764} & \makecell{38.2\%} & \makecell{0.409} & \makecell{16.5\%} & \makecell{0.523} & \makecell{21.31\%} \\
        & \makecell{SnapKV} & \makecell{0.872} & \makecell{9.6\%} & \makecell{0.285} & \makecell{16.22\%} & \makecell{0.51} & \makecell{16\%} & \makecell{0.5} & \makecell{1.79\%} & \makecell{0.2} & \makecell{1.8\%} & \makecell{0.44} & \makecell{2.79\% } & \makecell{0.293} & \makecell{8.2\%} & \makecell{0.443} & \makecell{8.06\%} \\ 
        & \makecell{Quest} & \makecell{0.866} & \makecell{20.4\%} & \makecell{0.24} & \makecell{16.02\%} & \makecell{0.48} & \makecell{15.8\%} & \makecell{0.6} & \makecell{20.98\%} & \makecell{0.333} & \makecell{25\%} & \makecell{0.5} & \makecell{39.9\%} & \makecell{0.561} & \makecell{13.8\%}& \makecell{0.511} & \makecell{21.7\%}\\
        & \makecell{ReasonCache} & \makecell{\underline{0.969}} & \makecell{21.5\%} & \makecell{\underline{0.36}} & \makecell{16.51\%} & \makecell{\underline{0.61}} & \makecell{15.9\%} & \makecell{\underline{0.7}} & \makecell{21.18\%} & \makecell{\underline{0.5}} & \makecell{23.2\%} & \makecell{\underline{0.918}} & \makecell{39.1\%} & \makecell{\underline{0.607}} & \makecell{15.54\%} & \makecell{\underline{0.665}} & \makecell{21.85\%}\\
        & \makecell{Dense} & \makecell{\textbf{0.982}} & \makecell{0\%} & \makecell{\textbf{0.375}} & \makecell{0\%} & \makecell{\textbf{0.62}} & \makecell{0\%} & \makecell{\textbf{0.733}} & \makecell{0\%} & \makecell{\textbf{0.533}} & \makecell{0\%} & \makecell{\textbf{0.927}} & \makecell{0\%} & \makecell{\textbf{0.631}} & \makecell{0\%} & \makecell{\textbf{0.686}} & \makecell{0\%} \\
        \midrule
        \multirow{5}{*}{QwQ 32B} 
        & \makecell{StreamingLLM} & \makecell{0.799} & \makecell{21.7\%} & \makecell{0.282} & \makecell{16.51\%} & \makecell{0.49} & \makecell{17.3\%} & \makecell{0.567} & \makecell{20.3\%} & \makecell{0.367} & \makecell{20.6\%} & \makecell{0.672} & \makecell{29.7\%} & \makecell{0.530} & \makecell{17.5\%}& \makecell{0.523} & \makecell{20.52\%}  \\
        & \makecell{SnapKV} & \makecell{0.756} & \makecell{6.2\%} & \makecell{0.32} & \makecell{15.7\%} & \makecell{0.61} & \makecell{15.5\%} & \makecell{0.1} & \makecell{1\% } & \makecell{0.233} & \makecell{1.43\% } & \makecell{0.176} & \makecell{1.55\% } & \makecell{0.146} & \makecell{4.27\% } & \makecell{0.334} & \makecell{6.52\%}\\
        & \makecell{Quest}& \makecell{0.915} & \makecell{20.8\%} & \makecell{0.36} & \makecell{15.82\%} & \makecell{0.56} & \makecell{17.8\%} & \makecell{0.666} & \makecell{25.1\%} & \makecell{0.533} & \makecell{23.5\%} & \makecell{0.506} & \makecell{28\%} & \makecell{0.535} & \makecell{18.1\%} & \makecell{0.582} & \makecell{21.30\%}\\
        & \makecell{ReasonCache}& \makecell{\textbf{0.972}} & \makecell{22.1\%} & \makecell{0.41} & \makecell{16.12\%} & \makecell{\textbf{0.652}} & \makecell{18.1\%} & \makecell{\underline{0.733}}& \makecell{23.41\%} & \makecell{\textbf{0.633}} & \makecell{23.1\%} & \makecell{\underline{0.872}} & \makecell{26.89\%} & \makecell{\underline{0.581}} & \makecell{18.12\%} & \makecell{\underline{0.693}} & \makecell{21.12\%}\\
        & \makecell{Dense} & \makecell{\underline{0.968}} & \makecell{0\%} & \makecell{\textbf{0.45}} & \makecell{0\%} & \makecell{\underline{0.648}} & \makecell{0\%} & \makecell{\textbf{0.766}} & \makecell{0\%} & \makecell{\textbf{0.633}} & \makecell{0\%} & \makecell{\textbf{0.894}} & \makecell{0\%} & \makecell{\textbf{0.611}} & \makecell{0\%}& \makecell{\textbf{0.71}} & \makecell{0\%} \\
        \midrule
        \multirow{5}{*}{\makecell{Phi 4\\ reasoning\\ plus}}
        & \makecell{StreamingLLM} & \makecell{0.812} & \makecell{14.8\%} & \makecell{0.278} & \makecell{14.8\%} & \makecell{0.45} & \makecell{13.4\%} & \makecell{0.567} & \makecell{16.6\%} & \makecell{0.367} & \makecell{17\%} & \makecell{0.522} & \makecell{23\%} & \makecell{0.333} & \makecell{18.9\%} & \makecell{0.476} & \makecell{16.93\%}\\
        & \makecell{SnapKV} & \makecell{0.793} & \makecell{11.5\%} & \makecell{0.292} & \makecell{15.5\%} & \makecell{0.47} & \makecell{13.7\%} & \makecell{0.2} & \makecell{2.56\% } & \makecell{0.166} & \makecell{3\% } & \makecell{0.146} & \makecell{4.76\% } & \makecell{0.253} & \makecell{7.29\% }& \makecell{0.331} & \makecell{8.33\%} \\
        & \makecell{Quest} & \makecell{0.854} & \makecell{15.3\%} & \makecell{0.29} & \makecell{15.75\%} & \makecell{0.47} & \makecell{13.6\%} & \makecell{0.567} & \makecell{18.5\%} & \makecell{0.433} & \makecell{17.7\%} & \makecell{0.518} & \makecell{21\%} & \makecell{0.312} & \makecell{17.8\%}& \makecell{0.478} & \makecell{17.09\%}\\
        & \makecell{ReasonCache} & \makecell{\textbf{0.945}} & \makecell{16.8\%} & \makecell{\underline{0.4}} & \makecell{16.2\%} & \makecell{\underline{0.61}} & \makecell{14.2\%} & \makecell{\textbf{0.8}} & \makecell{16.4\%} & \makecell{\underline{0.633}} & \makecell{17.2\%} & \makecell{\textbf{0.903}} & \makecell{21.41\%} & \makecell{\underline{0.508}} & \makecell{17.6\%} & \makecell{\underline{0.686}} & \makecell{17.12\%}\\
        & \makecell{Dense} & \makecell{\underline{0.933}} & \makecell{0\%} & \makecell{\textbf{0.43}} & \makecell{0\%} & \makecell{\textbf{0.635}} & \makecell{0\%} & \makecell{\textbf{0.8}} & \makecell{0\%} & \makecell{\textbf{0.677}} & \makecell{0\%} & \makecell{\underline{0.893}} & \makecell{0\%} & \makecell{\textbf{0.520}} & \makecell{0\%}& \makecell{\textbf{0.698}} & \makecell{0\%} \\
        \bottomrule
    \end{tabular}%
    }
    \label{table2}
    \vspace{-4mm}
  \end{table*}
 
      \noindent \textbf{Comparison with Other KV Cache Management Methods:}
      Table \ref{table2} presents a comprehensive comparison of various KV cache management methods across different models and benchmarks.     
    %  \update{To ensure a fair comparison across methods with different mechanisms, we adopt a unified definition of AMSR.
    %   For KV cache sharing, eviction and merging methods (e.g., ReasonCache, StreamingLLM and SnapKV), AMSR reflects the reduction in memory footprint.
    %   For selective KV loading methods (e.g., Quest), which optimize attention computation without releasing memory, AMSR measures the reduction in loaded KV cache data during inference.
    %   This unified metric allows us to evaluate the efficiency-accuracy trade-off across diverse strategies. }
    %   Additional experimental configurations for compared methods are provided in the Appendix~\ref{sec:exp_settings}.
      Under comparable Average Memory Saving Ratio, ReasonCache consistently outperforms other methods in terms of accuracy. 
      SnapKV, which is optimized for long prompts, demonstrates lower accuracy in mathematical and scientific reasoning tasks since it is less effective in long decoding scenarios.	
      StreamingLLM achieves moderate accuracy but suffers from information loss outside its sliding window.

      \subsection{Overhead Analysis}
\label{sec:overhead}
\begin{figure}[h!]
        \centering
        \begin{subfigure}[b]{0.3\columnwidth}
            \centering
            \includegraphics[width=\linewidth]{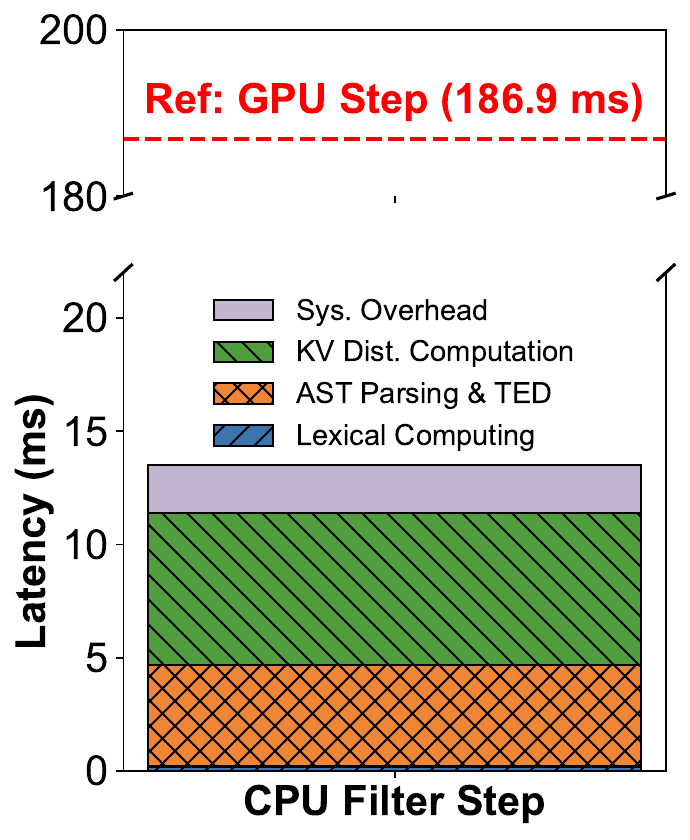}
            \caption{Overhead breakdown.}
            \label{table_compute_complexity}
        \end{subfigure}
        \hfill
        \begin{subfigure}[b]{0.68\columnwidth}
            \centering
            \includegraphics[width=\linewidth]{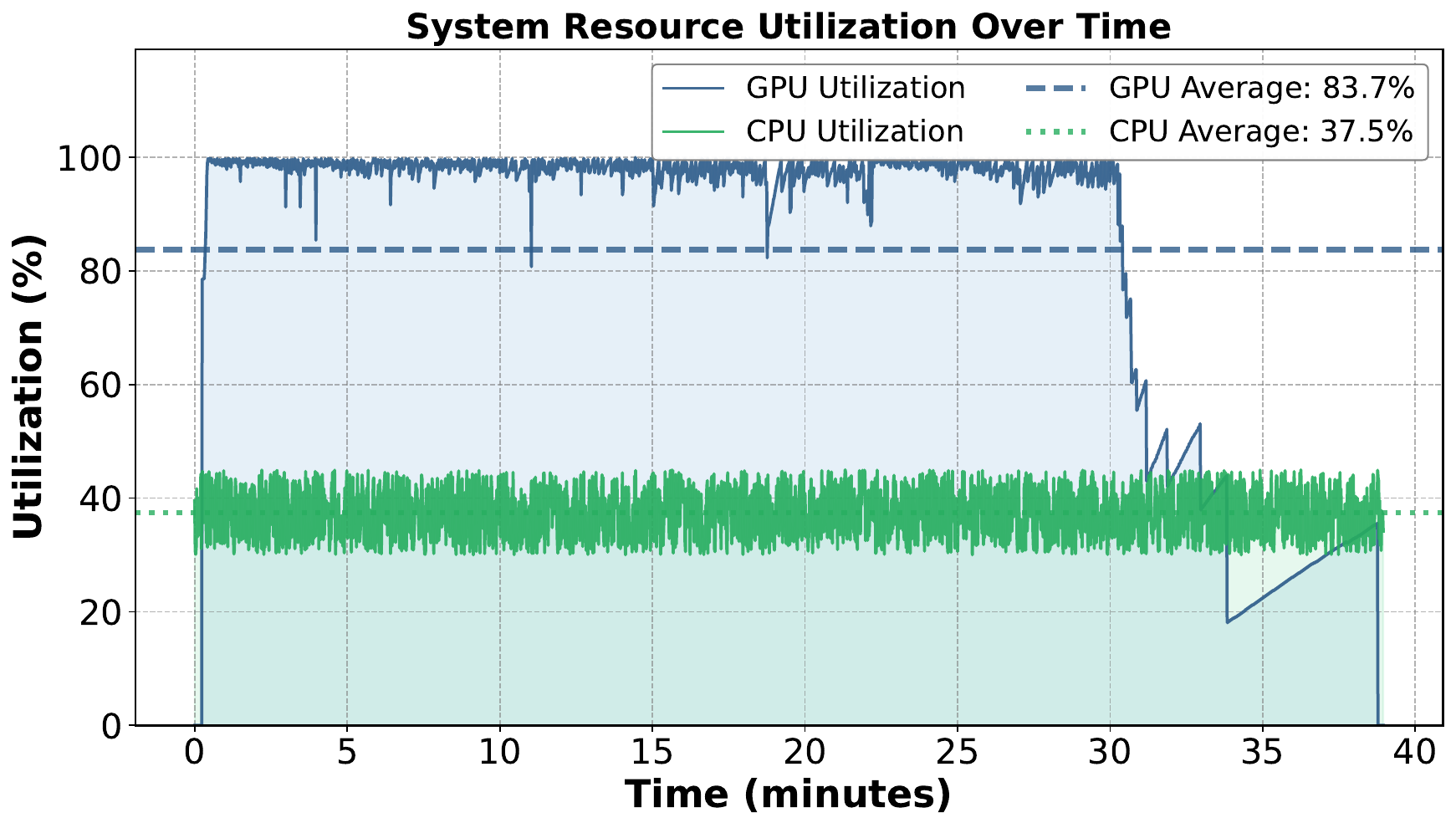}
            \caption{CPU/GPU utilization.}
            \label{util_data}
        \end{subfigure}
        \vspace{-2mm}
        \caption{Overhead breakdown per step and system utilization analysis.}
        \vspace{-5mm}
    \end{figure}

We conduct a detailed overhead analysis to quantify the system cost of ReasonCache, specifically focusing on the CPU-intensive filtering components.
Using the DeepSeek-R1-Distill-Qwen-32B model on the HumanEval dataset, we profile the average latency of each filtering stage, as reported in Figure~\ref{table_compute_complexity}.
The results show that ReasonCache introduces a modest average overhead of 13.5 ms per step.
In contrast, the model's average token generation time for a step is 186.9 ms.
Given that ReasonCache operates asynchronously, this CPU-side computation is completely masked by the GPU inference latency, ensuring zero impact on the critical path of token generation.

In addition to latency, we analyze the resource consumption to rule out potential bottlenecks.
Figure~\ref{util_data} presents the CPU and GPU utilization profiles under the same workload.
CPU utilization consistently fluctuates within the 35--45\% range, demonstrating that the filtering algorithm is efficient and far from saturating the CPU.
More importantly, GPU utilization is sustained at over 95\% during the initial 30-minute peak phase.
This high utilization confirms that our asynchronous offloading design effectively decouples CPU tasks from GPU inference, preventing pipeline stalls.
The subsequent drop in GPU utilization is due to the gradual completion of tasks, leading to reduced concurrency and smaller batch sizes in the final stages of the experiment.
    % \begin{table}[t!]
    %     \centering
    %     \scriptsize
    %     \caption{Overhead breakdown per step.}
    %     \setlength{\tabcolsep}{3mm}
    %     \resizebox{0.8\columnwidth}{!}{%
    %     \begin{tabular}{ccc}
    %         \toprule
    %         Component & Time (ms) & Ratio \\
    %         \midrule
    %         Lexical Filter & 0.2 & 1.5\% \\ 
    %         AST \& TED & 4.5 & 33.3\% \\
    %         KV Distance & 6.7 & 49.6\% \\
    %         System Overhead & 2.1 & 15.6\% \\
    %         \midrule
    %         \textbf{Total Filter Time} & \textbf{13.5} & \textbf{100\%} \\
    %         Ref: GPU Step & 186.9 & - \\
    %         \bottomrule
    %     \end{tabular}%
    %     }
    %     \label{table_compute_complexity}
    % \end{table}

      \subsection{Ablation Study}
      
       \noindent \textbf{Algorithm Design:}
      To evaluate the contribution of each component in our proposed algorithm, we conduct an ablation study to assess three distinct configurations:
      (1) Ours w/ only Cos, a version that utilizes only cosine similarity to measure the similarity between steps; 
      (2) Ours w/ only Cos + AST, which incorporates Abstract Syntax Tree (AST) analysis on top of cosine similarity to compare the structure of formal languages; 
      and (3) Ours, our full algorithm that combines cosine similarity, AST analysis, and Euclidean distance for a comprehensive similarity assessment.
      The results, presented in Table \ref{table3}, demonstrate that our full algorithm achieves the highest accuracy across all tasks. 
      However, this improved accuracy is accompanied by a decrease in throughput, which we attribute to the more granular filtering leading to less content sharing.
      
      \begin{table*}[h!]
        \centering
        \small
        \setlength{\tabcolsep}{0.6mm}
        \caption{Ablation study on different configurations of ReasonCache, comparing their accuracy and throughput (tokens/s).}% The highest accuracy for each benchmark is highlighted in bold.} 
        \resizebox{\textwidth}{!}{%
        \begin{tabular}{cccccccccccccc}
            \toprule
            & & \multicolumn{2}{c}{\textit{Code}} & \multicolumn{2}{c}{\textit{Multi-doc QA}} & \multicolumn{2}{c}{\textit{Summarization}} & \multicolumn{2}{c}{\textit{Math}} & \multicolumn{2}{c}{\textit{Science}} \\
            \cmidrule(lr){3-4} \cmidrule(lr){5-6} \cmidrule(lr){7-8} \cmidrule(lr){9-10}  \cmidrule(lr){11-12}
            \multirow{1}{*}{\makecell{Model}} & \multirow{1}{*}{\makecell{Setting}} & \multicolumn{2}{c}{\makecell{HumanEval}} & \multicolumn{2}{c}{\makecell{HotpotQA}} & \multicolumn{2}{c}{\makecell{TREC}} & \multicolumn{2}{c}{\makecell{MATH 500}} & \multicolumn{2}{c}{\makecell{GPQA Diamond}} &\makecell{Avg\\Acc} &\makecell{Avg \\throughput}\\
            % \cmidrule(lr){3-4} \cmidrule(lr){5-6} \cmidrule(lr){7-8} \cmidrule(lr){9-10} \cmidrule(lr){11-12} \cmidrule(lr){13-14} \cmidrule(lr){15-16}
            & & \makecell{Acc} & \makecell{throughput} & \makecell{Acc} & \makecell{throughput} & \makecell{Acc} & \makecell{throughput} & \makecell{Acc} & \makecell{throughput} & \makecell{Acc} & \makecell{throughput} \\
            \midrule
            \multirow{3}{*}{\makecell{DeepSeek R1\\ Distill\\Qwen 32B}}
            & \makecell{Ours w/ only Cos} & \makecell{0.902} & \makecell{277.99} & \makecell{0.31} & \makecell{572.21} & \makecell{0.535} & \makecell{436.21} & \makecell{0.78}  & \makecell{223.68} & \makecell{0.505} & \makecell{214.21} & \makecell{0.606} & \makecell{344.86}\\
            & \makecell{Ours w/ only Cos+AST} & \makecell{0.933} & \makecell{255.85} & \makecell{0.31} & \makecell{570.64} & \makecell{0.53} & \makecell{439.35} & \makecell{0.822}  & \makecell{210.06} & \makecell{0.556} & \makecell{198.23} & \makecell{0.63} & \makecell{334.83}\\
            & \makecell{Ours}& \makecell{\textbf{0.969}} & \makecell{246.01} & \makecell{\textbf{0.36}} & \makecell{542.43} & \makecell{\textbf{0.61}} & \makecell{416.84} & \makecell{\textbf{0.918}} & \makecell{194.5} & \makecell{\textbf{0.607}} & \makecell{176.21} & \makecell{\textbf{0.692}} & \makecell{315.198}\\
           \midrule
            \multirow{3}{*}{QwQ 32B} 
            & \makecell{Ours w/ only Cos} & \makecell{0.915} & \makecell{216.74} & \makecell{0.375} & \makecell{440.28} & \makecell{0.585} & \makecell{303.82} & \makecell{0.752}  & \makecell{181.25} & \makecell{0.48} & \makecell{142.12} & \makecell{0.621} & \makecell{256.84}\\
            & \makecell{Ours w/ only Cos+AST} & \makecell{0.939} & \makecell{201.26} & \makecell{0.375} & \makecell{444.39} & \makecell{0.59} & \makecell{301.39} & \makecell{0.8}  & \makecell{167.54} & \makecell{0.532} & \makecell{133.46} & \makecell{0.647} & \makecell{249.61}\\
            & \makecell{Ours} & \makecell{\textbf{0.972}} & \makecell{193.52} & \makecell{\textbf{0.41}} & \makecell{423.23} & \makecell{\textbf{0.652}} & \makecell{287.59} & \makecell{\textbf{0.872}}  & \makecell{152.31} & \makecell{\textbf{0.581}} & \makecell{125.67} & \makecell{\textbf{0.697}} & \makecell{236.464}\\
           \midrule
            \multirow{3}{*}{\makecell{Phi 4\\ reasoning\\ plus}}
            & \makecell{Ours w/ only Cos} & \makecell{0.884} & \makecell{375.55} & \makecell{0.365} & \makecell{728.21} & \makecell{0.54} & \makecell{578.92} & \makecell{0.792}  & \makecell{430.24} & \makecell{0.404} & \makecell{278.12} & \makecell{0.597} & \makecell{478.21}\\
            & \makecell{Ours w/ only Cos+AST} & \makecell{0.939} & \makecell{358.63} & \makecell{0.37} & \makecell{730.67} & \makecell{0.54} & \makecell{580.78} & \makecell{0.85}  & \makecell{412.32} & \makecell{0.449} & \makecell{252.52} & \makecell{0.63} & \makecell{466.98}\\
            & \makecell{Ours} & \makecell{\textbf{0.945}} & \makecell{338.33} & \makecell{\textbf{0.4}} & \makecell{695.88} & \makecell{\textbf{0.61}} & \makecell{556.84} & \makecell{\textbf{0.903}} & \makecell{384.21} & \makecell{\textbf{0.508}} & \makecell{243.22} & \makecell{\textbf{0.673}} & \makecell{443.696}\\
            \bottomrule
        \end{tabular}%
        }
        \label{table3}
        \vspace{-4mm}
      \end{table*}

      \begin{figure}[h!]
        \centering
        \begin{subfigure}[t]{0.49\linewidth}
            \includegraphics[width=\textwidth]{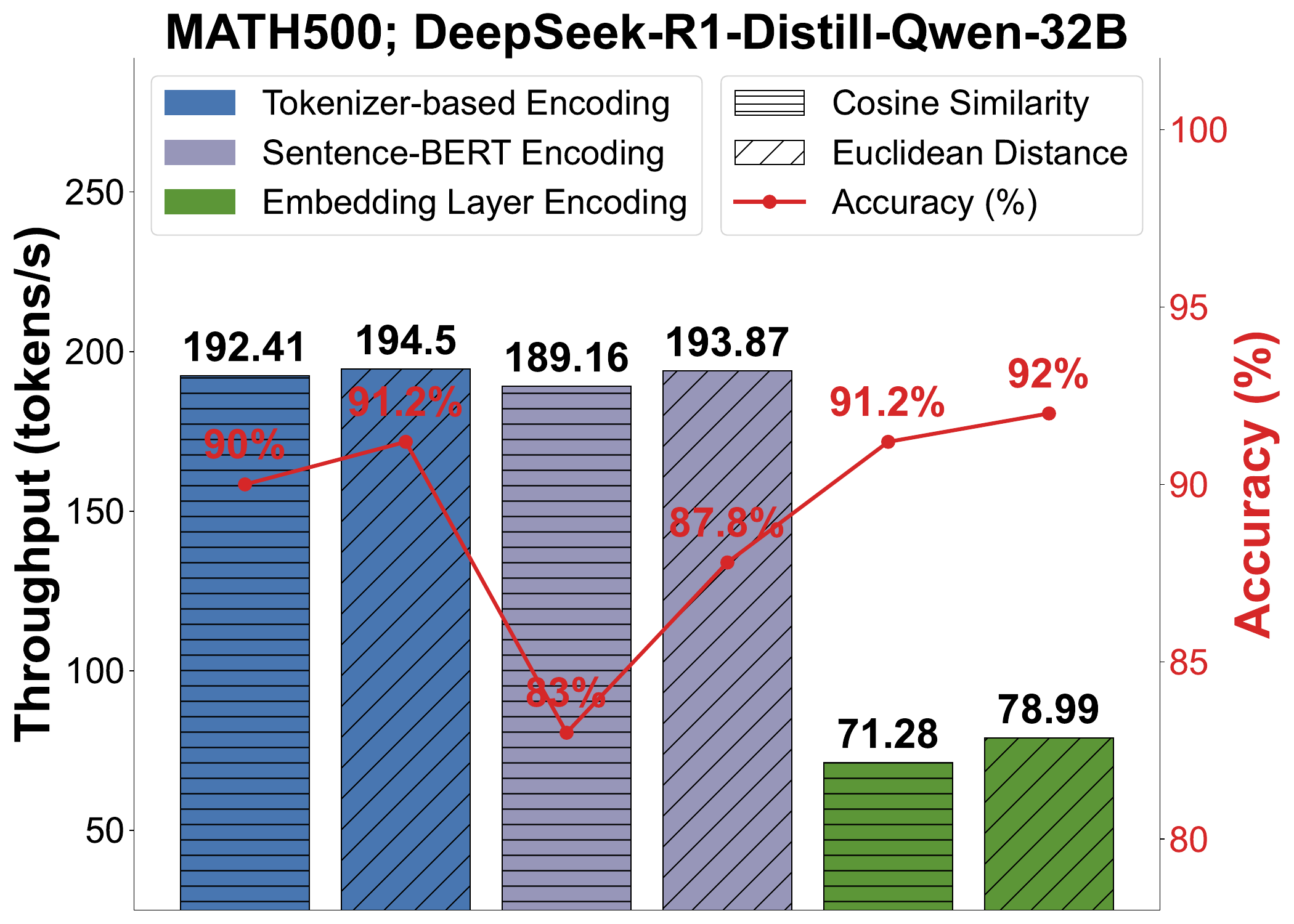}
        \end{subfigure}
        \hfill
        \begin{subfigure}[t]{0.49\linewidth}
            \includegraphics[width=\textwidth]{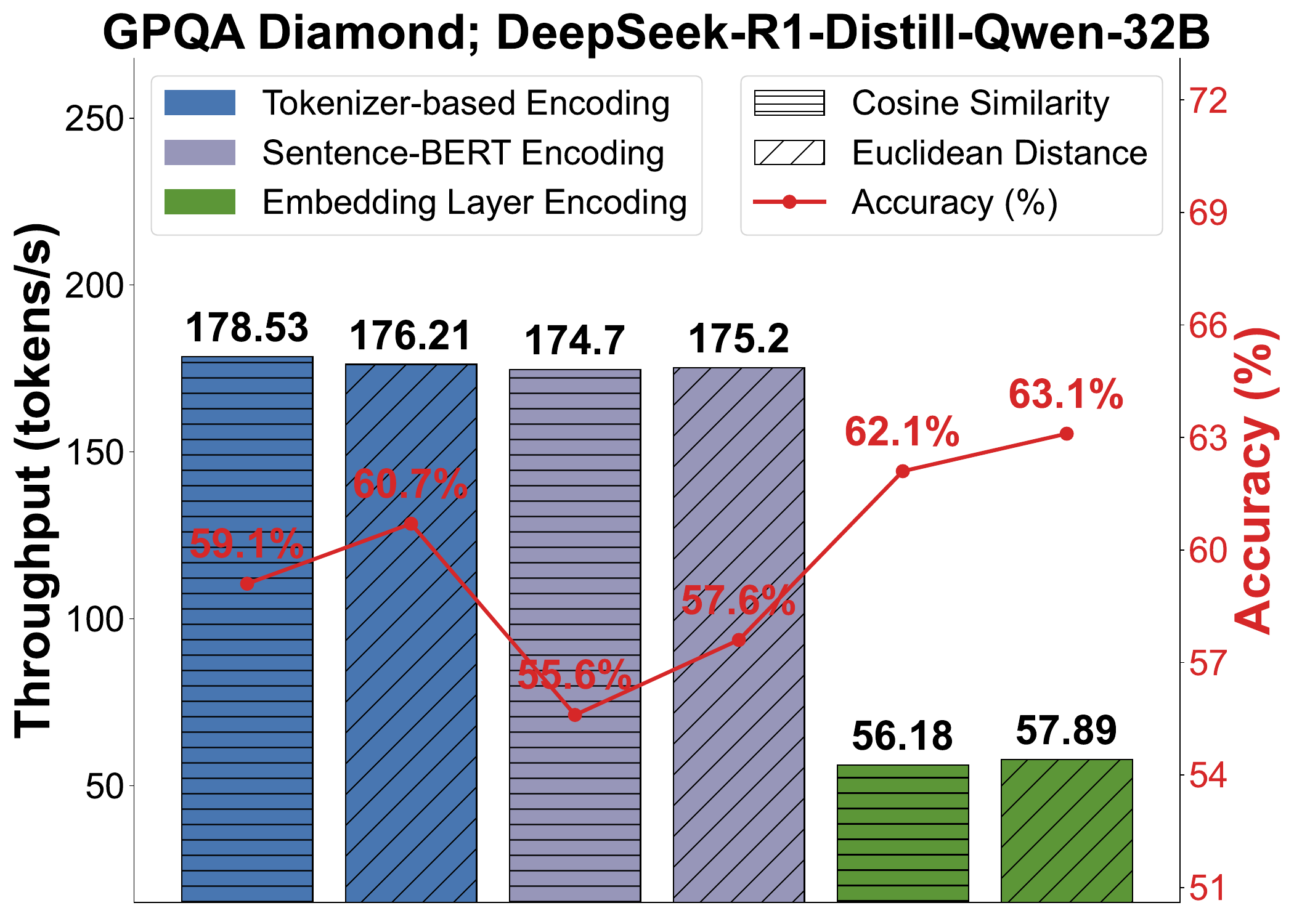}
        \end{subfigure}
        \caption{Performance comparison of similarity measurement algorithms. Algorithms are categorized in step-level (distinguished by colors) and block-level (distinguished by line patterns).}
        \label{fig5}
      %   \vspace{-3mm}
    \end{figure}
      \noindent \textbf{Similarity Measurement Methods:} 
      Figure \ref{fig5} evaluates the effectiveness of different similarity measurement strategies, addressing the potential concern that simple lexical matching might miss semantic differences. 
      We specifically compare our lightweight tokenizer-based approach against semantic embedding models.
      We assess three encoding approaches for step-level filtering:
      (1) \textbf{Tokenizer-based Encoding}: Our default method, using sparse bag-of-words vectors from the model's native tokenizer.
      (2) \textbf{Sentence-BERT}~\cite{sentence-bert}: Using \textit{all-MiniLM-L6-v2} to generate 384-dimensional dense semantic embeddings. This introduces an explicit semantic validation step.
      (3) \textbf{Embedding Layer}: This approach uses the the model's native embedding layer, which, while potentially more accurate, incurs higher computational costs.
    
      Our results demonstrate that Tokenizer-based Encoding combined with Block-Level Euclidean Distance yields the highest accuracy.
      We attribute this to two factors:
      First, the KV cache states already encode high-level semantic and contextual information specific to the model~\cite{semanticlocating}. Calculating the Euclidean distance between these states (Stage 2) effectively acts as an intrinsic semantic verification, rendering external semantic checks redundant.
      Second, external models like Sentence-BERT do not perfectly align with the target LRM's internal representation.
      Thus, our two-stage design efficiently balances precision and throughputs.

      \begin{figure}[t] 
          \centering
          \begin{subfigure}[t]{0.49\linewidth}
              \includegraphics[width=\textwidth]{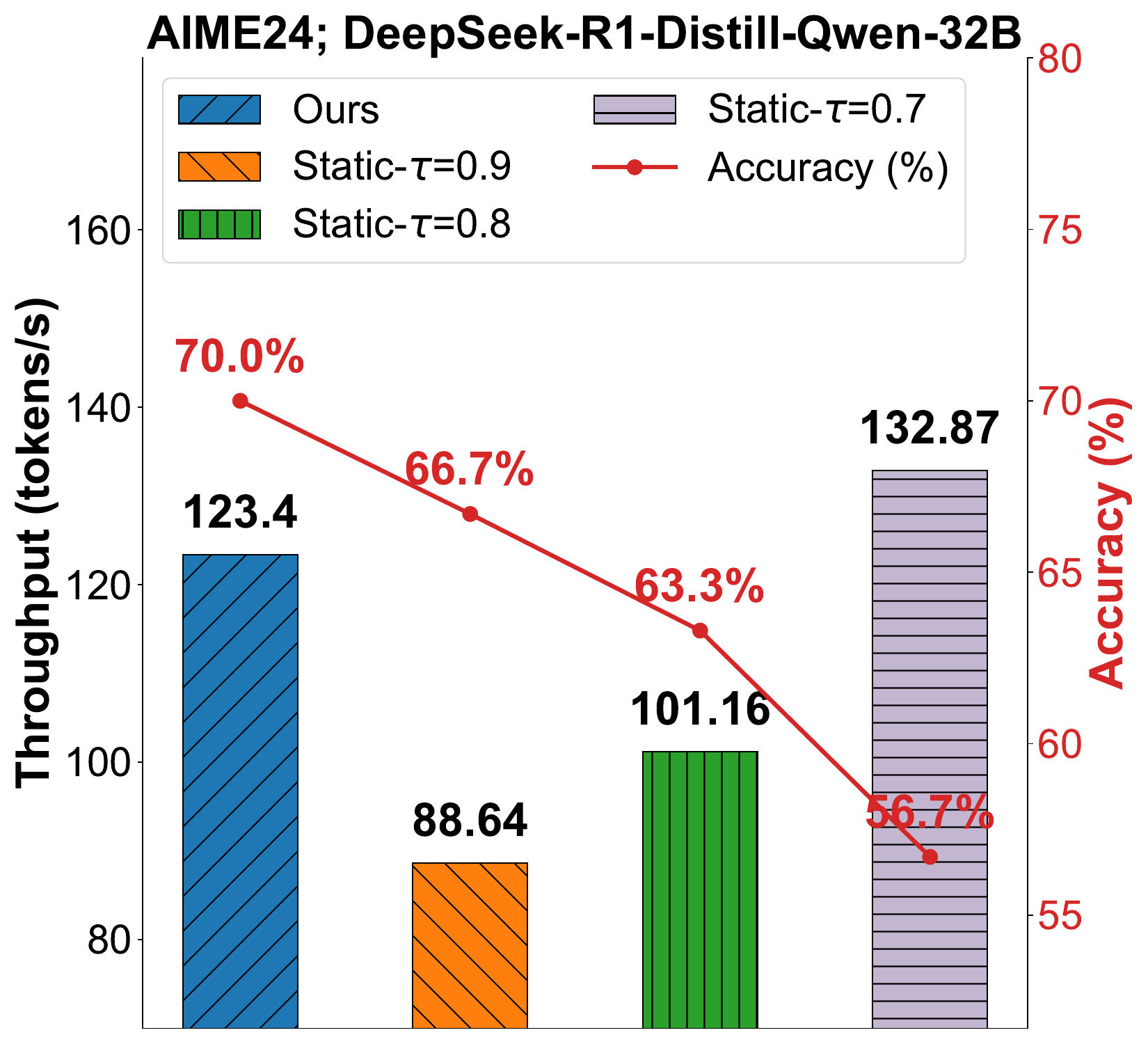}
          \end{subfigure}
          \hfill
          \begin{subfigure}[t]{0.49\linewidth}
              \includegraphics[width=\textwidth]{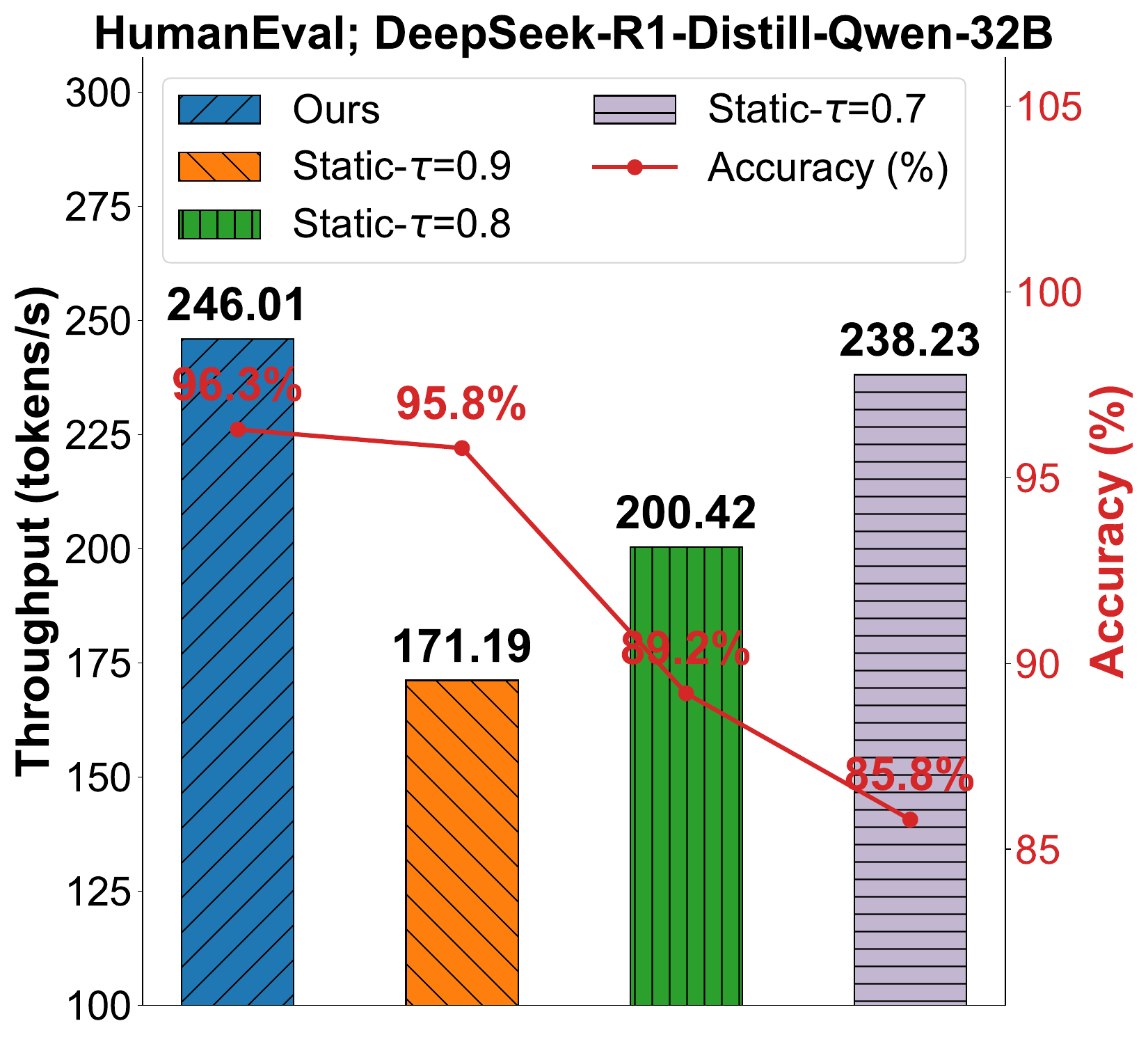}
          \end{subfigure}
          \caption{Comparison of dynamic and fixed thresholding.}
          \label{fig6}
          \vspace{-4mm}
        \end{figure}

        \begin{table}[h!]
          \centering
          \footnotesize
          \setlength{\tabcolsep}{0.7mm}
          \caption{Performance comparison of different percentile values (p) for Percentile-based Adaptive Thresholding (PAT).}%  For each benchmark, the highest and second-highest accuracies are highlighted in bold and underlined, respectively.}
          \resizebox{\columnwidth}{!}{%
          \begin{tabular}{clllllll}
              \toprule
              \makecell{Model}& \makecell{Threshold}  & \multicolumn{2}{c}{\makecell{HumanEval}}  & \multicolumn{2}{c}{\makecell{TREC}} & \multicolumn{2}{c}{\makecell{MATH 500}} \\
              & & \makecell{Acc} & \makecell{throughput} & \makecell{Acc} & \makecell{throughput} & \makecell{Acc} & \makecell{throughput}\\
              \midrule
              \multirow{4}{*}{\makecell{DeepSeek R1\\ Distill\\Qwen 32B}}
              & \makecell{PAT-p=20} & \makecell{\textbf{0.97}} & \makecell{192.12} & \makecell{0.6} & \makecell{322.2} & \makecell{\textbf{0.924}} & \makecell{121.89} \\ 
              & \makecell{PAT-p=50} & \makecell{0.957}& \makecell{212.34} & \makecell{\underline{0.605}} & \makecell{352.13} & \makecell{\underline{0.922}} & \makecell{149.21}\\
              & \makecell{PAT-p=80} & \makecell{\underline{0.969}} & \makecell{246.01}& \makecell{\textbf{0.61}} & \makecell{416.84} & \makecell{0.918} & \makecell{194.5} \\
              & \makecell{PAT-p=100} & \makecell{0.933} & \makecell{255.85}& \makecell{0.53} & \makecell{439.35} & \makecell{0.822} & \makecell{210.06}  \\
              \midrule
              \multirow{4}{*}{QwQ-32B} 
              & \makecell{PAT-p=20} & \makecell{0.97} & \makecell{132.89} & \makecell{0.645} & \makecell{215.27}& \makecell{\textbf{0.878}} & \makecell{101.23} \\ 
              & \makecell{PAT-p=50} & \makecell{\textbf{0.976}}& \makecell{158.35} & \makecell{\textbf{0.655}} & \makecell{248.28}& \makecell{0.868} & \makecell{123.37} \\
              & \makecell{PAT-p=80} & \makecell{\underline{0.972}} & \makecell{193.52}& \makecell{\underline{0.652}} & \makecell{287.59} & \makecell{\underline{0.872}} & \makecell{152.31}  \\
              & \makecell{PAT-p=100} & \makecell{0.939} & \makecell{201.26}& \makecell{0.59} & \makecell{301.39} & \makecell{0.8} & \makecell{167.54} \\
              \midrule
              \multirow{4}{*}{\makecell{Phi 4\\ reasoning\\ plus}}
              & \makecell{PAT-p=20} & \makecell{\textbf{0.951}} & \makecell{261.23} & \makecell{\textbf{0.635}} & \makecell{412.3}& \makecell{0.893} & \makecell{310.2}  \\ 
              & \makecell{PAT-p=50} & \makecell{\underline{0.948}}& \makecell{295.71} & \makecell{\underline{0.62}} & \makecell{478.21} & \makecell{\underline{0.896}} & \makecell{342.3} \\
              & \makecell{PAT-p=80} & \makecell{0.945} & \makecell{338.33}& \makecell{0.61} & \makecell{556.84} & \makecell{\textbf{0.903}} & \makecell{384.21} \\
              & \makecell{PAT-p=100} & \makecell{0.939} & \makecell{358.63}& \makecell{0.54} & \makecell{580.78}& \makecell{0.85} & \makecell{412.32}   \\
              \bottomrule
          \end{tabular}%
          }
          \label{table4}
          \vspace{-4mm}
      \end{table}
      
      \noindent \textbf{Impact of threshold:}
      This section evaluates the impact of different thresholding strategies. We first compare our dynamic thresholding mechanism with several fixed threshold approaches.
      As shown in Figure~\ref{fig6}, the dynamic threshold, $\tau$, consistently achieves better accuracy and satisfying throughput compared to all tested fixed thresholds.
      Next, we investigate the effect of varying the percentile parameter, $P$, within our Percentile-based Adaptive Thresholding (PAT) framework. 
      The results in Table~\ref{table4} indicate that setting $P$ to the 80th percentile yields an optimal balance between accuracy and throughput. 
      This finding remains consistent across various models and datasets, demonstrating the robustness of this parameter choice for achieving high overall performance.
      
      \section{Related Work} 
      Recent work in KV cache management leverages the inherent sparsity of attention patterns to estimate the token importance. 
      \textbf{Eviction-based methods} reduce memory footprint by permanently discarding non-critical tokens.
      Strategies like H2O~\cite{h2o}, SnapKV~\cite{snapkv}, and RaaS~\cite{raas} rely on importance scoring metrics.
      Other approaches, such as StreamingLLM~\cite{streaming-llm}, maintain a fixed-size cache by prioritizing attention sinks.
      Recent specialized methods for reasoning tasks include Lethe~\cite{lethe}, which introduces layer- and time-adaptive pruning and RLKV~\cite{RLKV} which adapts reinforcement learning for identifying reasoning-critical attention head.
      However, eviction-based methods face a fundamental limitation: discarded KV pairs cannot be recovered even if needed later~\cite{scbench}.
      \textbf{Selective KV loading} methods, in contrast, retain the full KV cache but skip non-critical tokens during attention computation. 
      Methods such as Quest~\cite{quest} exemplify this approach. These techniques reduce information loss but maintain the original memory footprint of the KV cache. 
       \textbf{KV cache merging} methods combine less important tokens into more critical ones rather than discarding them entirely. 
       MiniCache~\cite{Minicache} proposes an intra layer merging algorithm with restoration mechanisms to reduce memory usage. 
       D2O~\cite{d2o} leverages cosine similarity between key states to select merge candidates and determine dynamic weights.
       However, the difficulty in accurately predicting which tokens are important for future text generation can lead to information loss, causing hallucinations~\cite{yang2024no}.
      
      \section{Conclusion}
      
      We present ReasonCache, a novel KV cache management approach that efficiently identifies and reuses similar KV cache blocks.
      Comprehensive evaluations show that ReasonCache achieves a peak throughput improvement of 89.2\% and an average gain of 40-60\%, while reducing KV cache memory usage by 17-21\%, all without compromising accuracy.
    %   Compared to existing KV cache management methods under comparable Average Memory Saving Ratio, 
    %   ReasonCache consistently delivers better accuracy-efficiency trade-offs. 
      
      \bibliographystyle{IEEEtran}
      \bibliography{references}
           
      \appendices

      \vspace{-3mm}
      \section{Proof of Theorem 1}
      \label{sec:proof}
      \textit{Let $d \in \mathbb{N}$ denote the hidden state dimension. For a given decoding step $t$, let $q_t \in \mathbb{R}^d$ be the query vector. 
      Let $K_t = [k_1, k_2, \dots, k_t]^\top \in \mathbb{R}^{t \times d}$ and $V_t = [v_1, v_2, \dots, v_t]^\top \in \mathbb{R}^{t \times d}$ be the key and value matrices, 
      respectively, where $k_j, v_j \in \mathbb{R}^d$ for $j = 1, \dots, t$. The unnormalized attention scores, denoted by $S_t \in \mathbb{R}^t$, 
      are given by: $S_t = \frac{q_t K_t^\top}{\sqrt{d}}$.}
  
      \textit{The attention weights $A_t \in \mathbb{R}^t$ are then computed by applying the softmax function to $S_t$: $A_t = \mathrm{softmax}\left(\frac{q_t K_t^\top}{\sqrt{d}}\right)$}.
      \textit{The output vector $o_t \in \mathbb{R}^d$ at decoding step $t$ is a weighted sum of the value vectors: $o_t = \sum_{j=1}^t (A_t)_j v_j$}.
      
      \textit{Suppose for some $j \in \{1, \ldots, t\}$, we replace $k_j$ with $k_j'$ such that $\|k_j - k_j'\|_2 < \varepsilon$, and $v_j$ with $v_j'$ such that $\|v_j - v_j'\|_2 < \delta$. 
      We will analyze the effect of these substitutions on the attention weights and the output $o_t$.}

      \noindent\textbf{1. Key Similarity $\implies$ Similar Attention Scores}\\
      Let $S_t$ be the vector of unnormalized attention scores with key matrix $K_t$, and let $S_t'$ be the vector of unnormalized attention scores when $k_j$ is replaced by $k_j'$ in $K_t$ (forming $K_t'$). The $i$-th component of $S_t$ is $(S_t)_i = \frac{q_t k_i^\top}{\sqrt{d}}$. Similarly, the $i$-th component of $S_t'$ is:
      $(S_t')_i = \begin{cases} 
      \frac{q_t k_i^\top}{\sqrt{d}}, & i \neq j \\ 
      \frac{q_t (k_j')^\top}{\sqrt{d}}, & i = j 
      \end{cases}$.
  
      The difference between $S_t$ and $S_t'$ is non-zero only at the $j$-th position. The magnitude of this difference is:
      {\small $$|(S_t)_j - (S_t')_j| = \left|\frac{q_t k_j^\top}{\sqrt{d}} - \frac{q_t (k_j')^\top}{\sqrt{d}}\right| = \frac{|q_t (k_j - k_j')^\top|}{\sqrt{d}}$$ }
      By the Cauchy-Schwarz inequality, $|q_t (k_j - k_j')^\top| \leq \|q_t\|_2 \|k_j - k_j'\|_2$. Given $\|k_j - k_j'\|_2 < \varepsilon$, we have:
      $$|(S_t)_j - (S_t')_j| < \frac{\|q_t\|_2 \varepsilon}{\sqrt{d}}$$
      For $i \neq j$, $(S_t)_i - (S_t')_i = 0$. Therefore, the $\ell_\infty$ norm of the difference between $S_t$ and $S_t'$ is:
      {\small $$\|S_t - S_t'\|_\infty = \max_{i} |(S_t)_i - (S_t')_i| = |(S_t)_j - (S_t')_j| < \frac{\|q_t\|_2 \varepsilon}{\sqrt{d}}$$ }
      
      Now, we analyze the effect on the attention weights $A_t = \mathrm{softmax}(S_t)$ and $A_t' = \mathrm{softmax}(S_t')$. The softmax function $\sigma: \mathbb{R}^t \to \mathbb{R}^t$ is Lipschitz continuous~\cite{function-lipschitzness,lipschitz}. Specifically, for any $x, y \in \mathbb{R}^t$, the following inequality holds:
      $$\|\sigma(x) - \sigma(y)\|_1 \leq \|x - y\|_\infty$$
      Applying this property to our case, we get:
      {\footnotesize
      $$\|A_t - A_t'\|_1 = \|\mathrm{softmax}(S_t) - \mathrm{softmax}(S_t')\|_1 \leq \|S_t - S_t'\|_\infty < \frac{\|q_t\|_2 \varepsilon}{\sqrt{d}}$$
      }
      This result demonstrates that a small perturbation in a key vector $k_j$ leads to a linearly bounded change in the $\ell_1$ norm of the attention weights, directly proportional to the magnitude of the perturbation $\varepsilon$ and the norm of the query vector, and inversely proportional to $\sqrt{d}$.

      \noindent\textbf{2. Value Similarity $\implies$ Small Output Perturbation}
      
      Let $o_t$ be the original output and $o_t'$ be the output when only the value vector $v_j$ is replaced by $v_j'$, while keeping the key matrix $K_t$ (and thus $A_t$) unchanged. The expressions for $o_t$ and $o_t'$ are:
      {\small
      $o_t = \sum_{i=1}^t (A_t)_i v_i ,  o_t' = \sum_{i \neq j} (A_t)_i v_i + (A_t)_j v_j'$
      }
      The difference between the two outputs is:
      $$\|o_t - o_t'\|_2 = \|(A_t)_j v_j - (A_t)_j v_j'\|_2 = \|(A_t)_j (v_j - v_j')\|_2$$
      Since $(A_t)_j$ is a scalar, we have:{\small $\|o_t - o_t'\|_2 = |(A_t)_j| \|v_j - v_j'\|_2$}
      
      Given that $A_t$ is a probability distribution, $(A_t)_j \geq 0$ and $\sum_{i=1}^t (A_t)_i = 1$, which implies $(A_t)_j \leq 1$. With $\|v_j - v_j'\|_2 < \delta$, we can write:
      $\|o_t - o_t'\|_2 < (A_t)_j \delta \leq \delta$
      
      This shows that if only a single value vector is perturbed, the change in the output $o_t$ is directly bounded by the magnitude of the perturbation $\delta$.
      
      More generally, if a set of value vectors indexed by $J \subseteq \{1, \dots, t\}$ are perturbed, i.e., $v_j \to v_j'$ for all $j \in J$ such that $\|v_j - v_j'\|_2 < \delta$ for all $j \in J$, then by the triangle inequality:
      {\small
      \begin{align*}
          \|o_t - o_t'\|_2 &= \left\|\sum_{j \in J} (A_t)_j (v_j - v_j')\right\|_2 \leq \sum_{j \in J} \|(A_t)_j (v_j - v_j')\|_2 \\
          &= \sum_{j \in J} (A_t)_j \|v_j - v_j'\|_2 < \sum_{j \in J} (A_t)_j \delta = \delta \sum_{j \in J} (A_t)_j
      \end{align*}
      }
      Since {\scriptsize $\sum_{j \in J} (A_t)_j \leq \sum_{j=1}^t (A_t)_j = 1$}, it follows that:
      {\scriptsize $\|o_t - o_t'\|_2 < \delta$}
      
      Thus, even with multiple value vector perturbations, the total change in the output is bounded by the maximum perturbation in any individual value vector.
      
      \noindent\textbf{3. Combined Effect on Output}
      
      If both $k_j$ and $v_j$ are replaced, the total change in $o_t$ can be decomposed as:
      {
      $\|o_t - o_t''\| \leq \|o_t - o_t'\| + \|o_t' - o_t''\|$
      }.

      where $o_t''$ is the output with both $k_j \to k_j'$ and $v_j \to v_j'$. The first term is bounded as above. The second term, due to the change in $A^t$ from $k_j \to k_j'$, is:
      {\scriptsize
      \[
      \|o_t' - o_t''\| = \left\|\sum_{i=1}^t (A_i^{t} - A_i^{t'}) v_i'\right\| \leq \sum_{i=1}^t |A_i^{t} - A_i^{t'}| \cdot \|v_i'\| \leq \|A_t - A_t'\|_1 \cdot \max_i \|v_i'\|
      \]
      }
      Assuming $\|v_i'\|$ is bounded (as is typical in practice), this term is $O(\varepsilon)$.
      
      \noindent\textbf{Conclusion.}\\
      The above analysis rigorously establishes that, for sufficiently small $\varepsilon$ and $\delta$, the perturbations in the attention weights and output are both tightly bounded. 

\end{document}